\documentclass[10pt,twocolumn,letterpaper]{article}
\usepackage[pagenumbers]{arxivstyle}

\usepackage{colortbl}
\usepackage{amssymb}
\usepackage{pifont}
\usepackage{booktabs,multirow}
\usepackage{makecell}
\usepackage{tabulary}
\usepackage{fontawesome5}
\usepackage{bbding}

\newlength\savewidth\newcommand\shline{\noalign{\global\savewidth\arrayrulewidth
  \global\arrayrulewidth 1pt}\hline\noalign{\global\arrayrulewidth\savewidth}}

\usepackage{amsmath}
\usepackage{capt-of}
\usepackage[dvipsnames]{xcolor}
\usepackage{xspace}
\usepackage{enumitem}
\usepackage{amsmath,amssymb,amsbsy,amsfonts,dsfont,pifont,bm,bbm,mathrsfs,mathtools,nicefrac}
\usepackage{algorithm,algpseudocode,listings}
\usepackage{booktabs,multirow,adjustbox,diagbox,threeparttable,tabularray}

\newcommand{\methodname}{Mimir}
\newcommand{\method}{\texttt{\methodname}\xspace}
\newcommand{\tfive}{T5\xspace}

\newcommand{\llama}{Phi3\xspace}

\newcommand{\tocite}[1]{{\color{red} [TO CITE]}}

\definecolor{CQColor}{rgb}{0.0,0.0,1.0} %

\definecolor{TSColor}{rgb}{0.5,0.0,0.8} %

\definecolor{CQRColor}{rgb}{1.0,0.0,1.0} %

\definecolor{gongbiaoblue}{rgb}{0.21,0.49,0.74}
\usepackage[pagebackref,breaklinks,colorlinks,allcolors=gongbiaoblue]{hyperref}
\usepackage{wrapfig}
\usepackage[capitalize]{cleveref}  %
\crefname{section}{Sec.}{Secs.}
\Crefname{section}{Section}{Sections}
\crefname{table}{Tab.}{Tabs.}
\Crefname{table}{Table}{Tables}
\crefname{figure}{Fig.}{Figs.}
\Crefname{figure}{Figure}{Figures}
\crefname{equation}{Eq.}{Eqs.}
\Crefname{equation}{Equation}{Equations}
\definecolor{baseColor}{rgb}{0.75,0.05,0.1}
\newcommand{\base}[1]{{\color{baseColor}#1}}
\definecolor{checkmarkColor}{rgb}{0.1,0.75,0.1}
\newcommand{\checkc}[1]{{\color{checkmarkColor}#1}}
\definecolor{demphcolor}{RGB}{144,144,144}

\renewcommand{\thefootnote}{\fnsymbol{footnote}} %

\title{\methodname: Improving Video Diffusion Models for Precise Text Understanding}

\author{Shuai Tan$^{1}$\footnotemark[1]\>\,\footnotemark[2]\>\,, Biao Gong$^{1}$\footnotemark[1]\>\,\footnotemark[3]\>\,, Yutong Feng$^{2}$, Kecheng Zheng$^{1}$, Dandan Zheng$^{1}$, Shuwei Shi$^{1}$,\\ Yujun Shen$^{1}$, Jingdong Chen$^{1}$, Ming Yang$^{1}$\\[5pt]
{$^1$Ant Group\ \ $^2$Tsinghua University}\\
}

\begin{document}
\twocolumn[{
\maketitle
\begin{center}
    \vspace{-12pt}
    \includegraphics[width=0.95\linewidth]{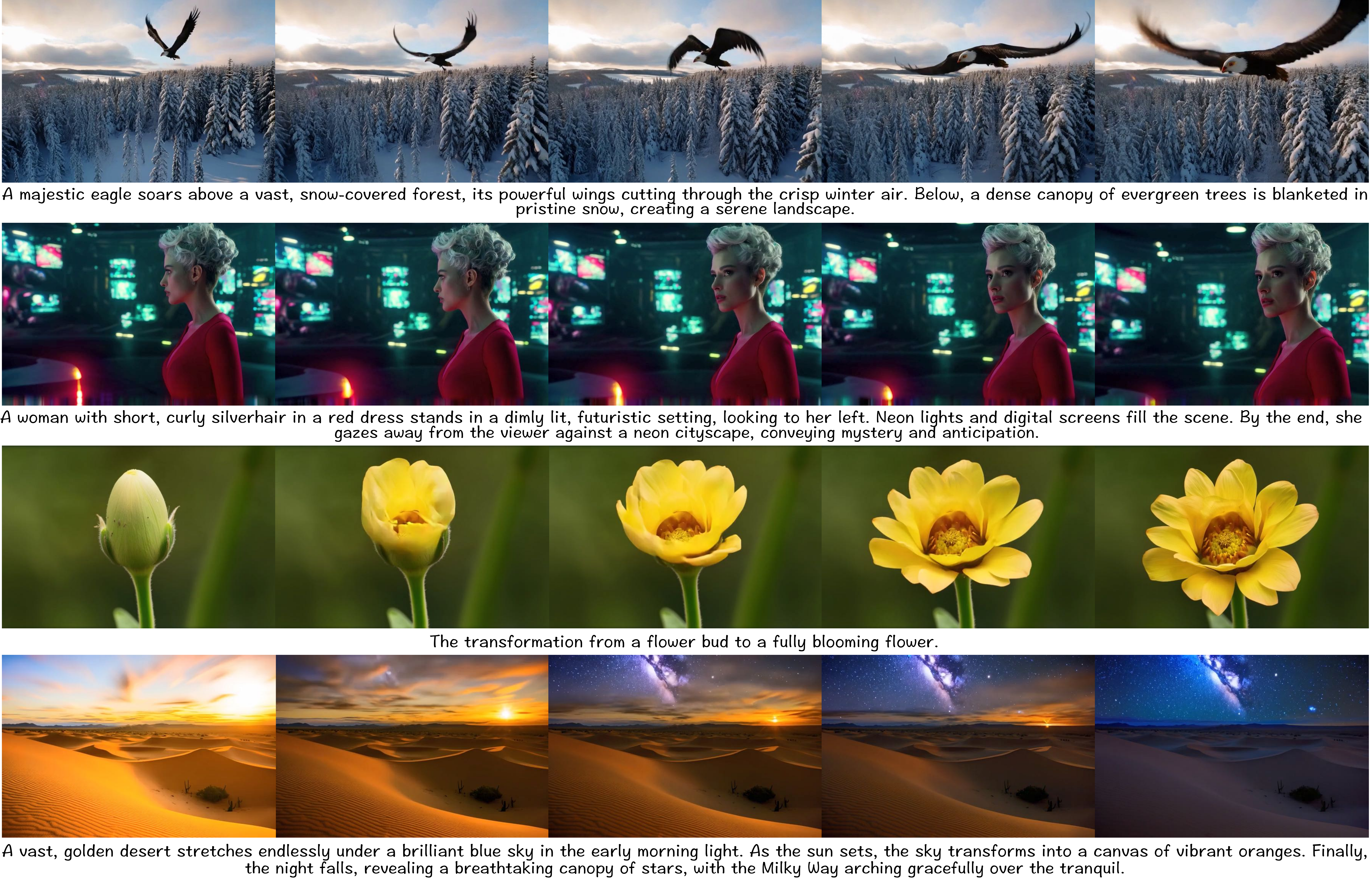}
        \vspace{-8pt}
    \captionof{figure}{
        Samples generated by \method. Our model demonstrates a powerful spatiotemporal imagination for input text prompts, \textit{e.g.}, \textit{(row-3)} physically accurate petals, \textit{(row-4)} the desert with illumination harmonization, which closely match human cognition.
    }
    \vspace{2mm}
    \label{fig:teaser}
\end{center}
}]

\footnotetext[1]{Equal contribution. \footnotemark[2]Work done during internship at Ant Group.}
\footnotetext[3]{Project lead and corresponding author.}

\begin{abstract}

\vspace{-3mm}
\noindent Text serves as the key control signal in video generation due to its narrative nature.
To render text descriptions into video clips, current video diffusion models borrow features from text \textbf{encoders} yet struggle with limited text comprehension.
The recent success of large language models (LLMs) showcases the power of \textbf{decoder-only} transformers, which offers three clear benefits for text-to-video (T2V) generation, namely,
precise text understanding resulting from the superior scalability,
imagination beyond the input text enabled by next token prediction,
and flexibility to prioritize user interests through instruction tuning.
Nevertheless, the feature distribution gap emerging from the two different text modeling paradigms hinders the direct use of LLMs in established T2V models.
This work addresses this challenge with \method, an end-to-end training framework featuring a carefully tailored \textbf{token fuser} to harmonize the outputs from text encoders and LLMs.
Such a design allows the T2V model to fully leverage learned video priors while capitalizing on the text-related capability of LLMs.
Extensive quantitative and qualitative results demonstrate the effectiveness of \method in generating high-quality videos with excellent text comprehension, especially when processing short captions and managing shifting motions.
Project page: \href{https://lucaria-academy.github.io/Mimir/}{https://lucaria-academy.github.io/Mimir/}

\end{abstract}
\vspace{-20pt}

\section{Introduction}
\label{sec:intro}

Language is the most natural and efficient way for human to convey perspectives and creative ideas after thousands of years of evolution~\cite{feng2024ranni, gong2024check}. 
To render text descriptions into video clips, current video diffusion models borrow features from text encoders yet struggle with limited text comprehension~\cite{cogvideo, stablediffusion2, blattmann2023stable}. 
Many diffusion based studies have explored powerful text encoders such as CLIP~\cite{clip} and T5~\cite{t5}, which still yield limited text understanding, particularly in video generation.
In fact, human-provided concise prompts cannot capture the vast spatiotemporal visual details in videos, such as the speed of a moving car or background changes along its route. This limitation has motivated researchers to explore semantic enhancement using large language models (LLMs), given their remarkable capabilities in text-related tasks~\cite{llama2, phi3}.

The recent success of LLMs showcases the power of decoder-only transformers, which offers three clear benefits for T2V generation.
\textit{Firstly}, it ensures precise text understanding which stems from terabytes of training data and the scalability of LLMs.
\textit{Secondly}, the capability for next token prediction allows the model to generate imaginative content that extends beyond the original input text, demonstrating creativity and contextual extrapolation.
\textit{Finally}, instruction tuning facilitates flexibility in prioritizing user interests, allowing the model to adapt its responses according to specific user directives.
Therefore, we aim to achieve the integration of heterogeneous (\textit{i.e.}, encoder and decoder-only) LLMs to improve video diffusion models especially for precise text understanding.

Achieving such integration is challenging due to the inherent volatility of decoder-only language models, \textit{i.e.}, these models prioritize predicting future tokens over representing the current text~\cite{llama2,ma2024exploring, xie2024sana}, thereby leading to the feature distribution gap and hindering the direct use of LLMs in established T2V models.
A promising approach involves fine-tuning a decoder-only model to function as an encoder~\cite{llama2}. Recent T2I~\cite{xie2024sana, feng2024ranni, ma2024exploring} have also explored various methods to enhance text prompt encoding.
However, we contend that these strategies constrain the full potential of decoder-only LLMs, particularly regarding their reasoning capabilities through next token prediction.

In this paper, we introduce \method which is an end-to-end training framework featuring a carefully tailored \textbf{Token Fuser} to harmonize the outputs from text encoders and decoder-only language models. Such a design allows \method to fully leverage learned video priors while capitalizing on the text-related capabilities of LLMs.
Specifically, the token fuser consists of two components. (1) It achieves non-destructive fusion by using Zero-Conv layers to merge the encoder tokens with all query and answer tokens generated by the decoder-only model.
This integration fully takes advantage of the LLM's capacity for reasoning.
(2) The proposed semantic stabilizer, which employs learnable parameters to stabilize fluctuating text features (\textit{e.g.}, features from different answers like `\textit{old car}', `\textit{dilapidated machine}', and `\textit{speeding car}', which are detailed in Sec.~\ref{sec:tokenfuser} and Sec.~\ref{sec:visual}).

\begin{figure}[t]
  \centering
    \includegraphics[width=1\linewidth]{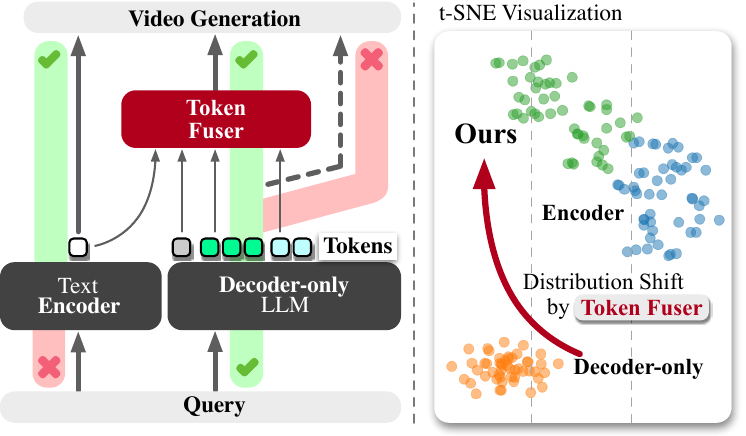}
    \caption{\textbf{The core idea of \method.}
    Text Encoder is well suited for fine-tuning pre-trained T2V models~(\checkc{\small \checkmark}), however it struggles with limited text comprehension~(\base{\tiny \XSolidBrush}). 
    In contrast, Decoder-only LLM excels at precise text understanding~(\checkc{\small \checkmark}), but cannot be directly used in established video generation models since the feature distribution gap and the feature volatility~(\base{\tiny \XSolidBrush}) .
    Therefore, we propose the token fuser in \method to harmonize multiple tokens, achieving precise text understanding~(\checkc{\small \checkmark}) in T2V generation~(\checkc{\small \checkmark}).}
    \label{fig:difference}
    \vspace{-4mm}
\end{figure}

In summary, as shown in Fig.~\ref{fig:difference}, \method integrates the decoder-only language model and ViT-style text encoder, trained within the diffusion framework to achieve precise text understanding~(\checkc{\small \checkmark}) in T2V generation~(\checkc{\small \checkmark}).
Extensive quantitative and qualitative (Fig.~\ref{fig:teaser}) results demonstrate the effectiveness of our approach in generating high-quality videos with excellent text comprehension, especially when processing short captions and managing shifting motions.

\begin{figure*}[t]
  \centering
    \includegraphics[width=1\linewidth]{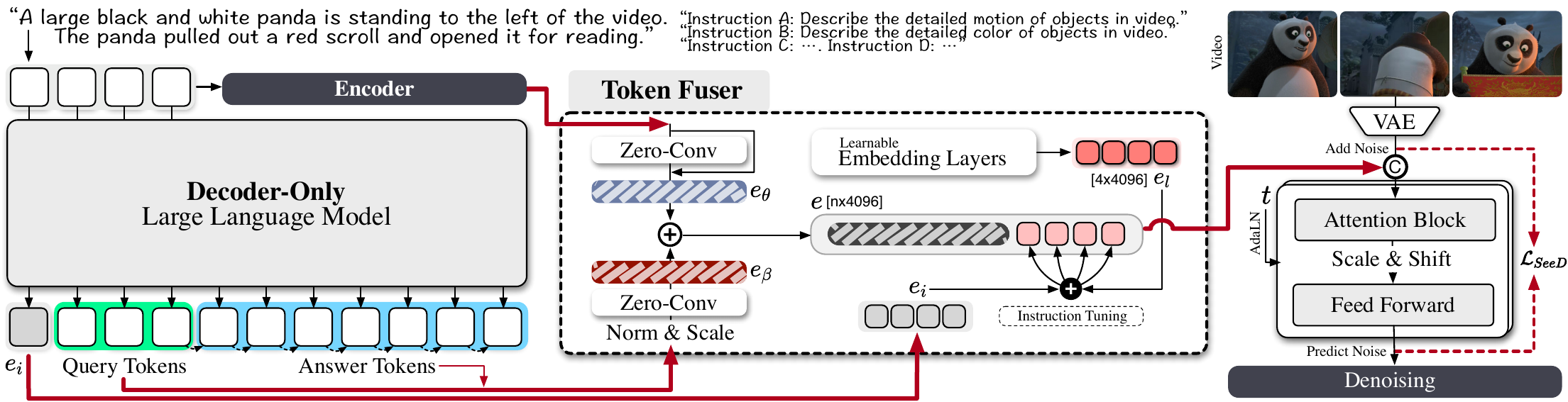}
    \caption{The framework of \method. Given a text prompt, we employ a text encoder and a decoder-only large language model to obtain $e_\theta$ and $e_\beta$. Additionally, we add an instruction prompt which, after processing by the decoder-only model, yields the corresponding instruction token $e_i$. See token details in Sec.~\ref{sec:tokens}. To prevent any convergence issue in training caused by the feature distribution gap of $e_\theta$ and $e_\beta$, the proposed token fuser first applies a normalization layer and a learnable scale to $e_\beta$. It then uses Zero-Conv to preserve the original semantic space in the early of training. These modified tokens are then summed to produce $e \in \mathbb{R}^{n\times4096}$. Meanwhile, we initialize four learnable tokens $e_l$, which are added to $e_i$ to stabilize divergent semantic features. Finally, the token fuser concatenates $e$ and $e_s$ to generate videos.}
    \label{fig:framework}
    \vspace{-2mm}
\end{figure*}

\section{Methodology}
\label{sec:method}

In this section, we first present the preliminaries of diffusion models in Section~\ref{sec:preliminary} and describe the different types of tokens in Section~\ref{sec:tokens}. Then, we introduce the details of the token fuser in Section~\ref{sec:tokenfuser}, which consists of two components: Non-Destructive Fusion and Semantic Stabilizer.

\subsection{Preliminary}
\label{sec:preliminary}

To lower the high training and inference costs of running diffusion models directly in pixel space, most diffusion-based models now follow the approach introduced by Rombach et al.~\cite{rombach2022high} known as latent diffusion models. This method typically consists of three key components: (a) \textit{Perceptual Video Compression and Decompression}: To efficiently handle video data, a pre-trained visual encoder~\cite{esser2021taming} is used to map the input video into a latent representation $z$. A corresponding visual decoder is then employed to reconstruct the latent representation back into the pixel space, yielding the reconstructed video $\hat{x} = \mathcal{D}(\mathcal{E}(x))$. (b) \textit{Semantic Encoding}: The text encoder is utilized to encode a given prompt into the text feature, which serves as the controlling signal for the content of the generated video. (c) \textit{Diffusion Models in Latent Space}: To model the actual video distribution, diffusion models~\cite{ho2020denoising, sohl2015deep} are used to denoise a normally distributed noise, aiming to reconstruct realistic visual content. 

Recent works on video generation commonly apply Diffusion in Transformer to carry out the denoising process. This process simulates the reverse of a Markov chain with a length of $T$. To reverse the process in latent space, noise $\epsilon$ is added to $z$ to obtain a noise-corrupted latent $z_t$ as described in~\cite{rombach2022high}. Subsequently, a Vision Transformer $\epsilon_{\theta}$ is used to predict the noise from $z_t$, the text embedding $e_\theta$, and the timestamp $t \in \{1, ..., T\}$. The optimization objective for this process can be formulated as:
\vspace{-2mm}
\begin{equation}
\mathcal{L}=\mathbb{E}_{\mathcal{E}(x), \epsilon \in \mathcal{N}(0,1), e_\theta, t}\left[\left\|\epsilon-\epsilon_{\theta}\left(\boldsymbol{z}_{t}, e_\theta, t\right)\right\|_{2}^{2}\right],
\vspace{-1mm}
\end{equation}
where $e_\theta$ refers to the text embedding. After the reversed denoising stage, the predicted clean latent is fed into the VAE decoder $\mathcal{D}$ to reconstruct the predicted video.

\subsection{Patches and Tokens}
\label{sec:tokens}

\noindent \textbf{Video Tokens.} 
Following~\cite{rombach2022high, cogvideox}, the core of video token construction lies in compressing the original RGB-T video into latent space and segmenting each frame of the video.
Specifically, we represent a video as 
\vspace{-2mm}
\begin{equation}
x \in \mathbb{R}^{(N + 1) \times H \times W \times 3},
\vspace{-2mm}
\end{equation}
where $(N + 1)$ represents the number of frames, $H$ and $W$ represent the height and width of each frame, respectively. Then we employ a 3D causal VAE~\cite{guo2023animatediff} $\mathcal{E}$ to compress it to the video latents $z \in \mathbb{R}^{(n + 1) \times h \times w \times C} = \mathcal{E}(x)$. Following the common setting in 3D VAEs for LVDMs~\cite{chen2024od, opensora, zhao2024cv}, the temporal rate $N/n$ and spatial rate $H/h = W/w$ are set as 4 and 8, respectively. Subsequently, we patchify the video latents $z$ to generate visual token sequence $z_\text{vision} \in \mathbb{R}^{\frac{(n + 1)}{q} \times \frac{h}{p} \times \frac{w}{p} \times C}$ with the length $q \cdot p \cdot p$.

\noindent \textbf{Text Tokens.} 
We provide two types of text tokens.
(1) Using the text encoder $\tau_\theta$, such as T5~\cite{t5}, to capture stable word-level text tokens from the input prompt $\mathcal{T}$. The process of converting text to tokens is referred to $e_\theta = \tau_\theta(\mathcal{T})$.
(2) Using the decoder-only LLM $\tau_\beta$, such as Phi-3.5~\cite{phi3}, leveraging its detailed understanding and reasoning capabilities~\cite{xie2024sana}, to capture text tokens with fluctuations but richer semantics. To fully preserve the extensive semantic capabilities, we retain all query and answer tokens as the final decoder-only tokens $e_\beta = \tau_\beta(\mathcal{T})$.
Besides, we also feed decoder-only LLM with four instruction prompts to generate instruction tokens $e_i$.
In the following sections, we will explain how the combination of these video and text tokens trains the transformer based T2V diffusion model.

\begin{table*}[!t]
\setlength\tabcolsep{2pt}
\def\w{20pt} 
\caption{%
    Quantitative results on VBench~\cite{vbench}. The best and second results for each column are \textbf{bold} and \underline{underlined}, respectively.
}
\vspace{-8pt}
\centering\footnotesize
\begin{tabular}{l@{\extracolsep{10pt}}c@{\extracolsep{10pt}}c@{\extracolsep{10pt}}c@{\extracolsep{10pt}}c@{\extracolsep{10pt}}c@{\extracolsep{10pt}}c@{\extracolsep{10pt}}c@{\extracolsep{10pt}}c}
\shline
\multirow{2}{*}{\textbf{Method}}             & \textbf{Background}   & \textbf{Aesthetic} &    \textbf{Imaging}  & \textbf{Object} & \textbf{Multiple} & \textbf{Color} & \textbf{Spatial} & \textbf{Temporal} \\\vspace{-4mm}\\
 & \textbf{Consistency} & \textbf{Quality} & \textbf{Quality} & \textbf{Class}& \textbf{Objects}& \textbf{Consistency}  & \textbf{Relationship}& \textbf{Style} \\
    \shline
ModelscopeT2V~\cite{VideoFusion}     & 92.00\%          & 37.14\%                & 55.85\%          & 31.17\%                & {1.52\%}  & 63.20\%          & 8.26\%           & 14.52\%           \\
OpenSora~\cite{opensora}     & 97.20\%          & 58.57\%                & 63.38\%          & {90.79\%}       & 64.81\%          & 84.67\%          & 76.63\%          & 25.51\%           \\
OpenSoraPlan~\cite{opensoraplan}  & 97.50\%          & {59.40\%}       & { 57.79\%}    & 67.39\%                & 26.98\%          & 83.38\%          & 38.69\%          & 21.86\%         \\
CogVideoX-2B~\cite{cogvideox}       & 94.71\%          & 60.27\%                & 60.52\%          & 84.86\%                & { 65.70\%}    & 86.21\%          & 70.49\%          & 25.10\%         \\
CogVideoX-5B~\cite{cogvideox}       & 95.60\%          & 60.62\%                & 61.35\%          & 87.82\%                & 65.70\%          & { 84.17\%}    & 64.86\%          & 25.86\%            \\
\hline
\textbf{\method}     & \textbf{97.68\%} & { \textbf{62.92\%}} & \textbf{63.91\%} & { \textbf{92.87\%}} & \textbf{85.29\%} & \textbf{86.50\%} & \textbf{78.67\%} & \textbf{26.22\%} \\
\shline
\end{tabular}
  \label{tab:Evaluation_results}%
\end{table*}%

\begin{figure*}[t]
  \centering
    \includegraphics[width=1\linewidth]{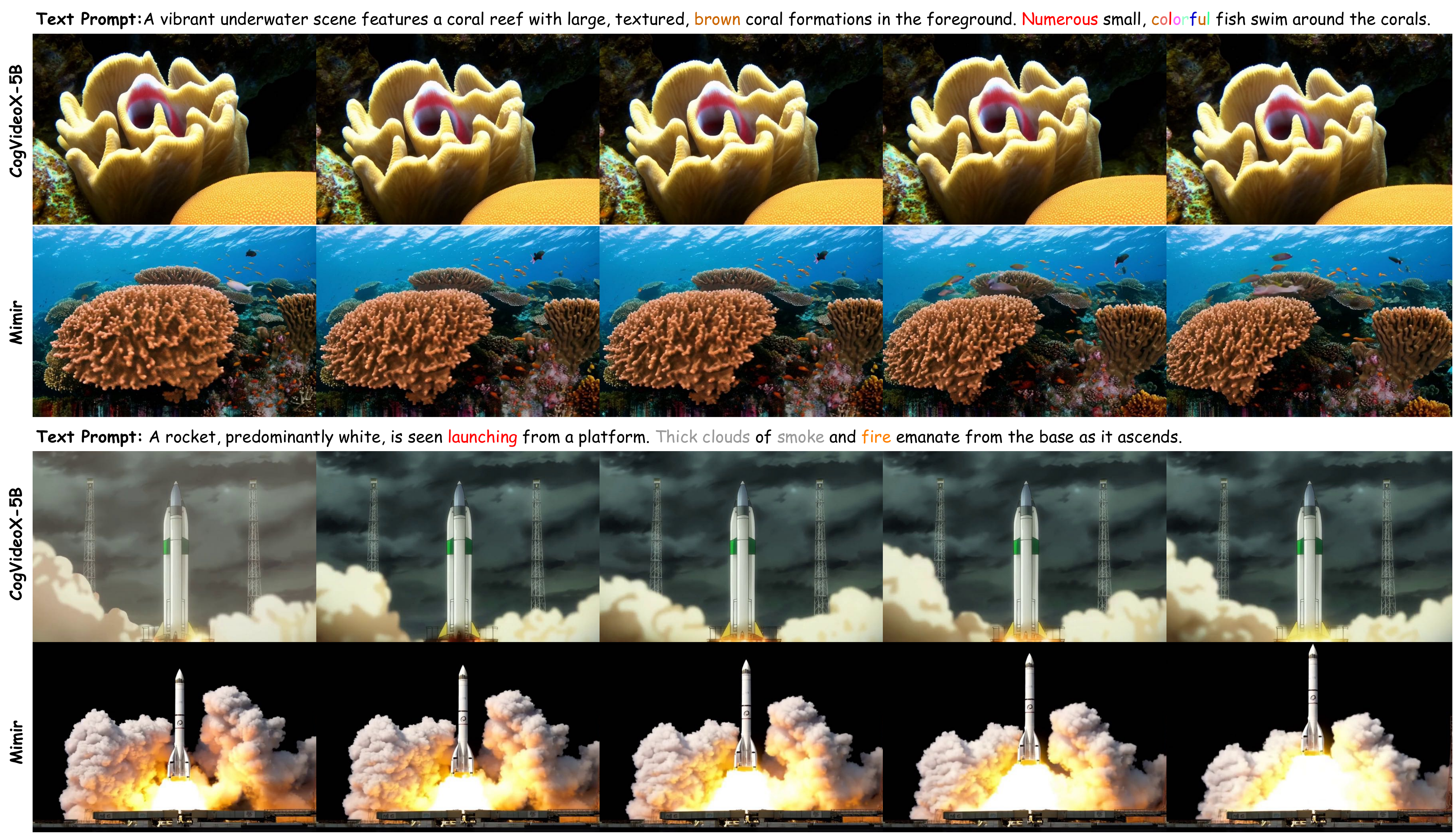}
    \vspace{-6mm}
    \caption{Comparison between CogVideoX-5B with \method in T2V, where \method generates the vivid stunning moment of rocket launch.}
    \vspace{-5mm}
    \label{fig:morecase}
\end{figure*}

\subsection{Token Fuser}
\label{sec:tokenfuser}

\noindent \textbf{Non-Destructive Fusion.}
As show in Fig.~\ref{fig:framework}, our method consists of two language branches (\textbf{\ie} encoder branch and decoder-only branch) and a vision transformer $\epsilon_\theta$. Both branches of language encode the input prompt $\mathcal{T}$ into tokens, and their sum is passed to $\epsilon_\theta$.
To mitigate the incompatibility between these embeddings, we implement two effective schemes: 
(1) \textbf{Normalization and Scaling}~\cite{xie2024sana}: We insert a normalization layer followed by a small learnable scale factor and bias directly after the decoder-only LLM. This step ensures that the two types of text tokens are brought to a similar scale, allowing them to be aligned in the fusion process. 
(2) \textbf{Zero Convolution Layer}: we introduce a zero-conv layer $\mathcal{Z}_{\beta}$ after the decoder-only token $e_\beta$. This ensures that the embedding $e_\beta$ starts as zero at the beginning of training.
Since our goal is to provide semantics from textual inputs to the Vision Transformer model $\epsilon_\theta$ from both the encoder and decoder-only branches, we need to balance their contributions throughout training.
Thus, we also insert a zero-conv layer $\mathcal{Z}_{\theta}$ after the encoder $\tau_\theta$ in a residual manner, ensuring that the embedding $e_\theta$ starts equal to the original tokens at the beginning. 
Compared to other commonly used adapters such as LoRA, zero-conv is lightweight and can smoothly achieve domain adaptation for textual or visual features.
Afterwards, we sum the embeddings as $e = e_\theta + \alpha \cdot e_\beta$ and feed $e$ into the Vision Transformer $\epsilon_\theta$, where $\alpha$ indicates the weight for decoder-only token:
\begin{equation}
e_\theta =  \tau_\theta(\mathcal{T}) + \mathcal{Z}_{\theta}(\tau_\theta(\mathcal{T})), \qquad 
e_\beta = \mathcal{Z}_{\beta}(\tau_\beta(\mathcal{T}))
\end{equation}

\begin{table*}[!t]
\setlength\tabcolsep{2pt}
\def\w{20pt} 
\caption{%
   User study results. The best and second results for each column are \textbf{bold} and \underline{underlined}, respectively.
}
\vspace{-8pt}
\centering\footnotesize
\begin{tabular}{l@{\extracolsep{10pt}}c@{\extracolsep{10pt}}c@{\extracolsep{10pt}}c@{\extracolsep{10pt}}c@{\extracolsep{10pt}}c@{\extracolsep{10pt}}c}
\shline
Method         &  ModelScopeT2V & OpenSora     & OpenSoraPlan & CogVideoX-2b& CogVideoX-5b      & \textbf{\method} \\
\shline
Instruction Following $\uparrow$ & {2.45\%}     & {52.15\%} & 27.75\%      & 63.50\%            & \underline{72.15\%} & \textbf{82.00\%} \\
Physics Simulation  $\uparrow$      & {3.50\%}     & 47.95\%          & 54.75\%      & 52.85\%            & \underline{57.30\%}          & \textbf{83.65\%}        \\
Visual Quality $\uparrow$ & {1.60\%}     & 49.20\%          & 41.50\%      & 54.80\%            &\underline{63.25\%}          & \textbf{89.65\%} \\

\shline
\end{tabular}
  \label{tab:user_study}%
\end{table*}%

\begin{figure*}[!t]
  \centering
    \includegraphics[width=1\linewidth]{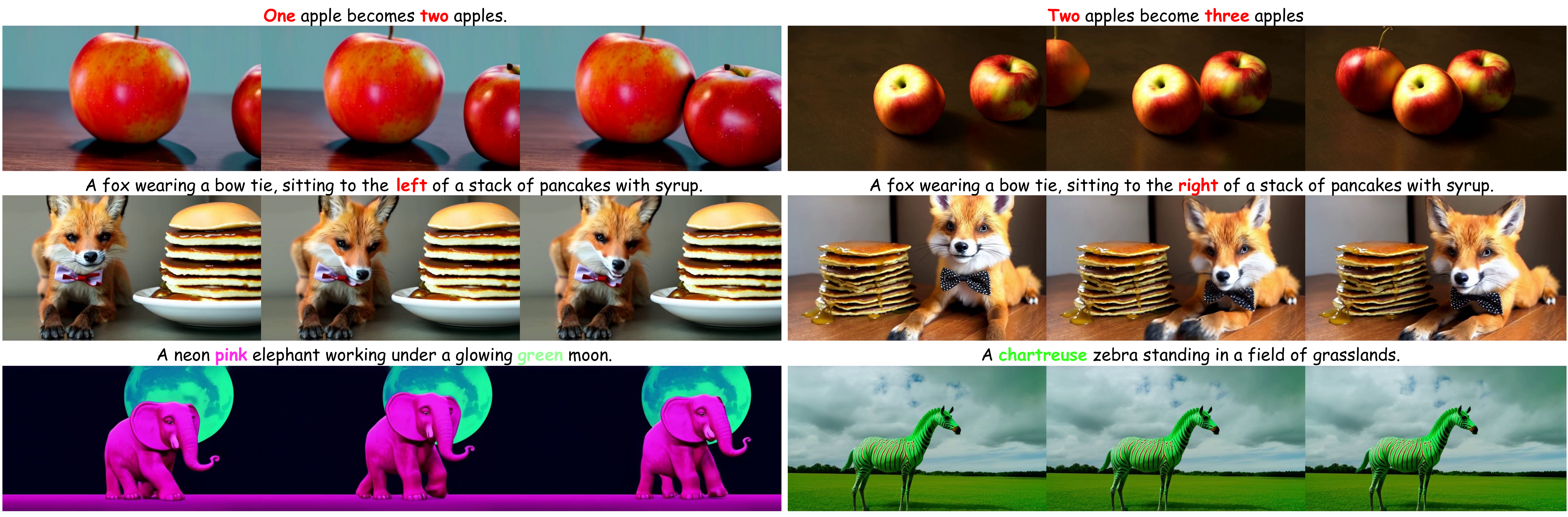}
    \vspace{-7mm}
    \caption{\method demonstrates spatial comprehension and imagination, \textbf{\textit{e.g.}}, quantities, spatial relationships, colors, etc.}
    \label{fig:num_location_color}
\end{figure*}

\begin{figure*}[!t]
  \centering
    \includegraphics[width=1\linewidth]{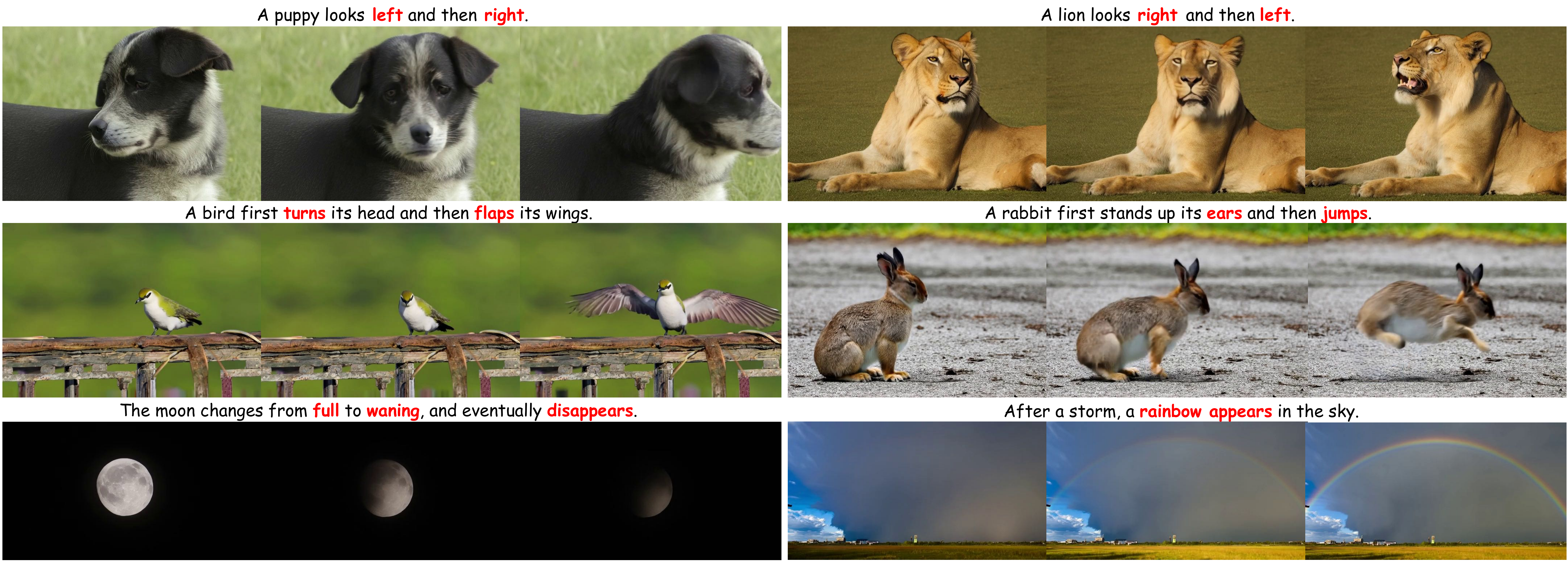}
    \vspace{-6mm}
    \caption{\method demonstrates temporal comprehension and imagination, \textbf{\textit{e.g.}}, direction, order of motion and appearance / disappearance.}
    \label{fig:time_case3}
\end{figure*}

\noindent \textbf{Semantic Stabilizer.} 
The Semantic Stabilizer serves two primary functions:
(1) To ensure the denoising model (\textit{i.e.}, the vision transformer) accurately captures the essential semantic elements in the prompt, such as \textit{object}, \textit{color}, \textit{motion}, and \textit{spatial relationships}.
(2) To stabilize the fluctuating textual features that emerge during next-token predictions, \textit{i.e.}, different descriptions of a `\textit{car}' such as `\textit{old car}' and `\textit{dilapidated machine}', which we analyze in detail in Sec.~\ref{sec:visual}.
Specifically, we begin by generating instruction tokens $e_i$ based on four pre-defined, attribute-specific instructions (\textit{e.g., Describe the detailed objects in the video}). 
Next, we initialize four learnable tokens $e_l$ with the same shape, designed as bridges to align with the visual space, resulting in our final semantic token $e_s = e_i + e_l$. We then concatenate $e_s$ along the sequence dimension to the previously defined token $e$. 
Finally, we concatenate the text-based token embeddings with the video embeddings and feed them together into the diffusion process.
To train the model, we minimize the diffusion loss, reducing the discrepancy between the predicted noise and the ground-truth noise during optimization. The overall loss is
\vspace{-2mm}
\begin{equation}
\mathcal{L}_{SeeD}=\mathbb{E}_{\mathcal{E}(x), \epsilon \in \mathcal{N}(0,1), \mathcal{T}, t}\left[\left\|\epsilon-\epsilon_{\theta}\left(\boldsymbol{z}_{t}, e \oplus e_s, t\right)\right\|_{2}^{2}\right],
\vspace{-2mm}
\end{equation}
where $\oplus$ refers to concatenation operation.

\section{Experiments}
\label{sec:exp}
In this section, we comprehensively evaluate our method and provide a detailed analysis of the reasons behind the effectiveness of our improvements, as well as the advantages of our approach in video generation performance.

\subsection{Text-to-Video Generation}
\noindent \textbf{Experimental Setup.} 
In the setting of LLMs, we select Phi-3.5~\cite{phi3} mini-instruct version as our decoder-only LLM to achieve a balance between computational efficiency and performance. 
In the setting of diffusion models, we implement v-prediction~\cite{salimans2022progressive} and zero SNR~\cite{lin2024common}, following the noise schedule established in LDM~\cite{rombach2022high}. We collect 500,000 high-quality video clips to train the \method model. We compare our approach against publicly accessible top-performing text-to-video models, including ModelscopeT2V~\cite{VideoFusion}, OpenSora~\cite{opensora}, OpenSoraPlan~\cite{opensoraplan}, CogvideoX-2B~\cite{cogvideox}, and CogvideoX-5B~\cite{cogvideox}. To evaluate the text-to-video generation, we employ several metrics from VBench~\cite{vbench}: \textit{Background Consistency} to assess temporal quality, \textit{Aesthetic Quality} and \textit{Imaging Quality} for frame-wise evaluation, as well as \textit{Object Class}, \textit{Multiple Objects}, \textit{Color Consistency}, \textit{Spatial Relationship}, and \textit{Temporal Style} for semantic understanding.

\begin{table*}[!t]
\setlength\tabcolsep{2pt}
\def\w{20pt} 
\caption{%
   Ablation study results. The best and second results for each column are \textbf{bold} and \underline{underlined}, respectively.
}
\vspace{-8pt}
\centering\footnotesize
\begin{tabular}{l@{\extracolsep{10pt}}c@{\extracolsep{10pt}}c@{\extracolsep{10pt}}c@{\extracolsep{10pt}}c@{\extracolsep{10pt}}c@{\extracolsep{10pt}}c@{\extracolsep{10pt}}c@{\extracolsep{10pt}}c}
\shline
\multirow{2}{*}{\textbf{Method}}             & \textbf{Background}   & \textbf{Aesthetic} &    \textbf{Imaging}  & \textbf{Object} & \textbf{Multiple} & \textbf{Color} & \textbf{Spatial} & \textbf{Temporal} \\\vspace{-4mm}\\
 & \textbf{Consistency} & \textbf{Quality} & \textbf{Quality} & \textbf{Class}& \textbf{Objects}& \textbf{Consistency}  & \textbf{Relationship}& \textbf{Style} \\
    \shline
Baseline    & 95.60\%       & 60.62\%          & 61.35\%       & 87.82\%         & {65.70\%} & 84.17\%       & 64.86\%         & \underline{ 25.86\%}          \\
 B+Decoder-only     &94.66\%       & 36.38\%          & 60.10\%       & {4.97\%} & 0.00\%           & 37.50\%       & 2.36\%          & 3.66\%           \\
B+Decoder-only+Norm  & 97.12\%       & {61.68\%} & \underline{ 62.52\%} & 85.50\%         & 65.24\%          & 84.85\%       & 59.28\%         & 25.26\%           \\
B+Decoder-only+Norm+SS       & 96.48\%       & 58.11\%          & 62.49\%       & 87.18\%         & \underline{ 68.83\%}    & 85.21\%       & 67.86\%         & 24.33\%          \\
B+Decoder-only+ZeroConv      & 97.20\%       & 61.21\%          & 62.99\%       & \underline{ 92.03\%}   & \underline{ 84.98\%}    & \underline{ 86.21\%} & 69.17\%         & 25.03\%       \\
B+Decoder-only+ZeroConv+SS      & \underline{ 97.33\%} & \underline{ 62.14\%}    & \underline{ 63.02\%} & 91.21\%         & 84.47\%          & 86.43\%       & \underline{ 70.16\%}   & 23.68\%            \\
\hline
\textbf{\method}     & \textbf{97.68\%} & { \textbf{62.92\%}} & \textbf{63.91\%} & { \textbf{92.87\%}} & \textbf{85.29\%} & \textbf{86.50\%} & \textbf{78.67\%} & \textbf{26.22\%} \\
\shline
\end{tabular}
  \label{tab:ablation}%
\end{table*}%

\begin{figure*}[!t]
  \centering
    \includegraphics[width=1\linewidth]{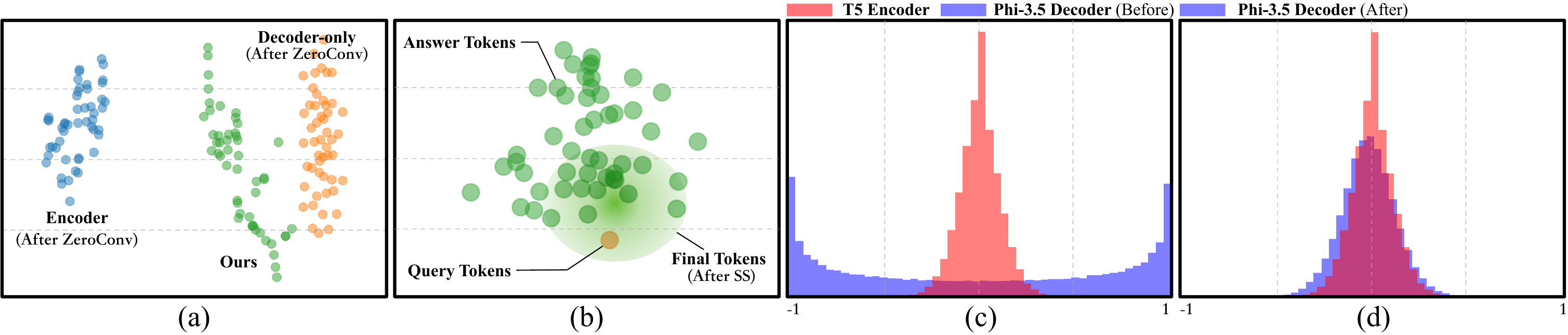}
    \caption{\textbf{Visualization by t-SNE:} (a) Given 50 prompts, we obtain the corresponding tokens using Encoder branch, Decoder-only branch and their sum, i.e., \method. (b) We feed one prompt into Decoder-only branch for 50 times to generate 50 query tokens, answer tokens and final tokens. \textbf{Differences in feature distribution:} (c) The original distribution of T5 encoder and Phi-3.5 Decoder. (d) The distribution of T5 encoder and Phi-3.5 Decoder after normalization across different value ranges.}
    \label{fig:hist}
    \vspace{-4mm}
\end{figure*}

\begin{figure*}[t]
  \centering
    \includegraphics[width=1\linewidth]{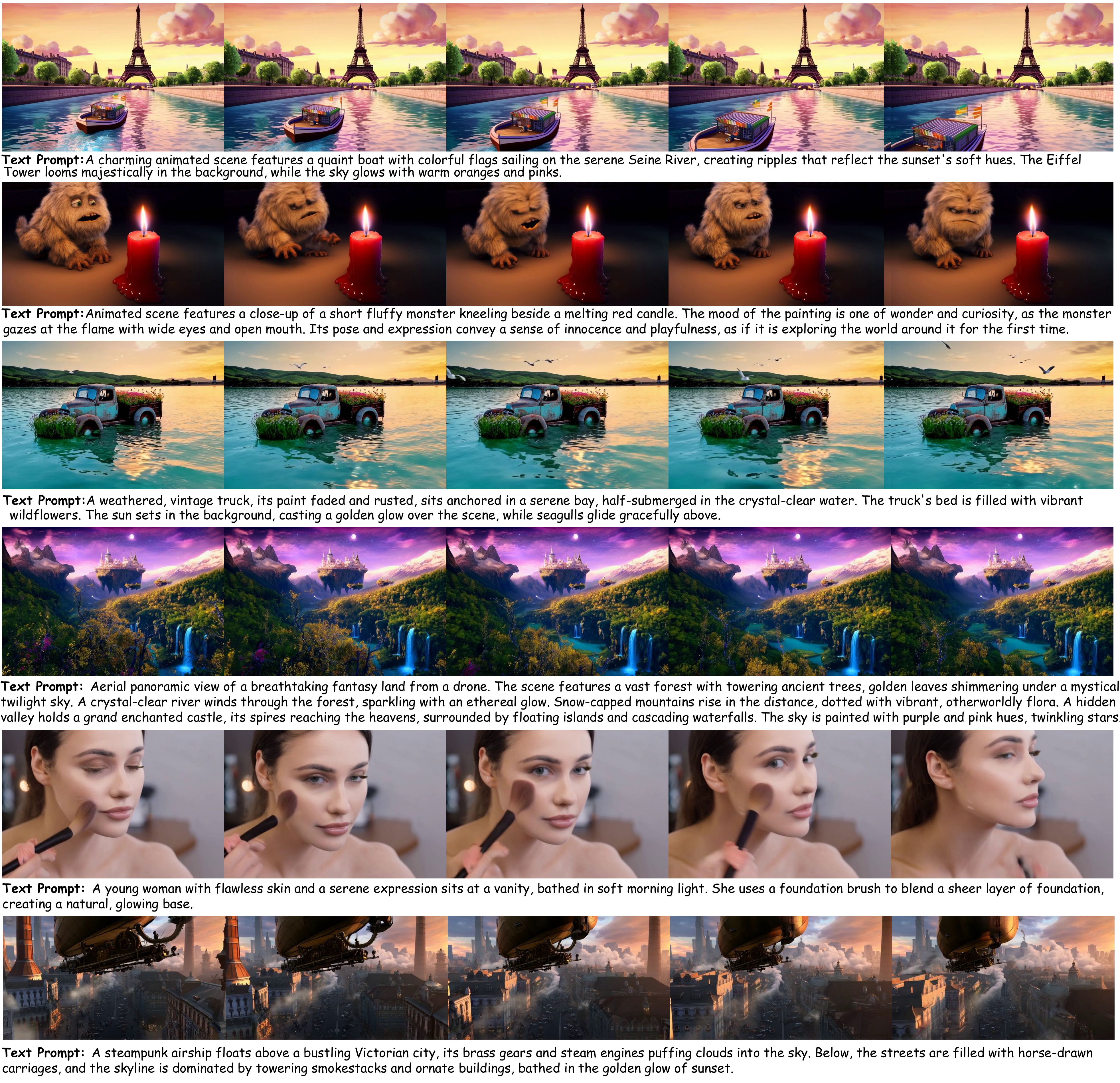}
    \vspace{-7mm}
    \caption{We present more cases generated by \method.}
    \label{fig:funcase}
    \vspace{-4mm}
\end{figure*}

\noindent \textbf{Quantitative Evaluation.} 
Tab.~\ref{tab:Evaluation_results} presents the evaluation results. \method outperforms existing approaches across all metrics. Notably, it shows significant improvements in the \textit{Multiple Objects} and \textit{Spatial Relationship} metrics. These results demonstrate that, with the assistance of LLMs, the video generation model achieves a marked performance enhancement compared to models that rely solely on the T5 encoder for semantic modeling.

\noindent \textbf{Qualitative Evaluation.} 
Fig.~\ref{fig:morecase} shows the comparison results between \method and the state-of-the-art method. With the support of the decoder-only Phi-3.5, \method is able to understand the input text prompt precisely, such as color, multiple objects, and quantities. Additionally, \method produces videos with high quality, showcasing its superior generative performance.

\noindent \textbf{User Study.} 
To evaluate the quality of \method and the SOTAs from a human perspective, we conducted a blind user study with 10 participants. We randomly select 20 prompts and feed them into each compared method and \method, resulting in a total of 120 video clips. Each participant is shown two videos generated by different methods for the same prompt and asked to choose which one performed better in terms of \textit{Instruction Following}, \textit{Physics Simulation}, and \textit{Visual Quality}. This process is repeated $C^6_2$ times. The results in Tab.~\ref{tab:user_study} show \method superior performance across all aspects.

\subsection{Ablation Studies} 
\label{sec:ablation}
\paragraph{Key Component}
To assess the effectiveness of the components in \method, we conduct an ablation study in a progressive manner. Specifically, the experiments are arranged as follows: (1) Baseline: Only T5 is used as the text encoder, with all other LLM components removed. (2) B+Decoder-only: The encoder token and Decoder-only token are directly combined. (3) B+Decoder-only+Norm: The Decoder-only token undergoes normalization before being combined with the encoder token. (4) B+Decoder-only+Norm+SS: The Semantic Stabilizer is added on top of (3). (5) B+Decoder-only+ZeroConv: The Encoder and Decoder-only tokens are fused using the Zero Conv method. (6) B+Decoder-only+ZeroConv+SS:  The SS is added on top of (5). (7) \method: The complete model with all components. The experimental results, as shown in Tab.~\ref{tab:ablation}, reveal that the direct combination of encoder and decoder-only tokens in (2) leads to model collapse due to the differences between the two token types. Normalization in (3) alleviates this issue, but semantic errors persist. With the addition of the Semantic Stabilizer in (4), a preliminary understanding of semantics is achieved. In (5), the Zero Conv method smoothly combines the two tokens. \method, with the aid of all modules, achieves the best performance.

\noindent \textbf{Spatial Comprehension and Imagination.} 
Based on the design of token fuser, our method accurately comprehends complex prompts, such as quantities, spatial relationships, and colors. For each aspect, we provide our method with 2 interesting prompts, as illustrated in Fig.~\ref{fig:num_location_color}.

\noindent \textbf{Temporal Comprehension and Imagination.} 
Another crucial aspect of the text-to-video task is temporal comprehension. This means that the generated video should not only meet requirements such as quantities, spatial relationships, and colors, as shown in Fig.~\ref{fig:num_location_color}, but also maintain the coherence and order between frames according to the prompt's instructions. Therefore, we further provide our method with several temporally related prompts. As shown in Fig.~\ref{fig:time_case3}, \method accurately understands the temporal relationships in the instructions, such as sequences from left to right or right to left. Moreover, when multiple actions are involved, our method comprehends the order of actions and generates the corresponding video.

\subsection{Visualization and Analysis}
\label{sec:visual}
Due to the different optimization functions of the encoder and the decoder-only model, there is a significant gap between their latent spaces. This gap can increase the difficulty of training, and may even lead to training collapse. To address this issue, we propose two solutions in Sec.~\ref{sec:method}: (1) adding a normalization layer and a learnable scale, and (2) incorporating a zero convolution layer. 

For the first solution, we randomly sample several prompts and encode them using encoder and decoder-only model, resulting in the corresponding encoder tokens and decoder-only tokens. We then input the decoder-only tokens into the normalization layer to obtain the normalized decoder-only tokens. Subsequently, we count the number of tokens within each numerical range. As visualized in Fig.~\ref{fig:hist} (c), the original encoder tokens' values are concentrated between -0.5 and 0.5, while decoder-only model has a much wider range that exceeds the -1 to 1 limits. After normalization in Fig.~\ref{fig:hist} (d), the magnitudes of the decoder-only tokens align with those of the encoder tokens, thereby reducing the training difficulty through this adjustment.

For the second solution, we use t-SNE~\cite{van2008visualizing} to reveal the distribution gap between encoder and decoder-only tokens, which is shown in Fig.~\ref{fig:hist} (a). 
The zero convolution prevents the direct summation of features with different distributions, allowing for a gradual integration of both types of tokens in the visual transformer during training.

Moreover, we explore the feature fluctuations of the same prompt in both text encoder and decoder-only language model.
Specifically, we randomly sample one prompt and encode it with decoder-only for 50 times, resulting in corresponding query tokens and answer tokens. The results are shown in Fig.~\ref{fig:hist}(b). 
We observe that the query tokens produce identical embeddings when encoding the same prompt multiple times, represented as a single point. In contrast, the results of answer tokens demonstrate the generative ability and the inherent volatility of decoder-only models, as encoding the same prompt yields a broader range of features. 
This phenomenon arises from the powerful reasoning abilities of decoder-only language models.
Specifically, when prompting LLMs to describe a `car', even with identical input text, the responses may vary (\textit{e.g.}, `\textit{old car}', `\textit{dilapidated machine}'), leading to the feature fluctuations.
However, completely eliminating these fluctuations would undermine the inherent strengths of decoder-only LLMs. Therefore, we use stable stabilizer to actively limit fluctuations and perform adaptive distribution adjustments for stable training.

\section{Related Work}
\label{sec:related_work}

\paragraph{Text-to-Video Generation.}
Video diffusion models~\cite{ho2022video} trains image and video jointly using 3D U-Net architecture and text conditions to handle the additional temporal dimension. 
Additionally, PYoCo~\cite{ge2023preserve} explores finetuning a pretrained image diffusion model with video data as a practical solution for the video synthesis task. It utilizes a noise prior and a pre-trained eDiff-I~\cite{balaji2022ediff} model for generating videos.
Subsequently, SVD~\cite{blattmann2023stable} pretrains a UNet-based image generation model~\cite{rombach2022high} and then add a temporal layer for video generation. 
Recently, inspired by the success of transformer-based model in text-to-image task, some works~\cite{opensora,cogvideox, cogvideo} utilize a diffusion transformer architecture to tackle challenges in long-duration and high-resolution video generation. 
However, current methods simply use CLIP or \tfive as text encoder, which limit the text understanding. Therefore, we aim to integrate superior decoder-only LLMs (such as \llama) into video diffusion model for optimizing the generated results by leveraging their precise understanding and reasoning capabilities.

\noindent\textbf{Large Language Model for Diffusion Framework.} 
Language models play a crucial role in image / video generation, and even some works~\cite{yu2023magvit, yu2024spae, kondratyuk2023videopoet, yu2023language} use only LLMs to generate images or videos, which reveals the powerful capability of LLMs. However, most of current diffusion models have not fully utilized the advantages of LLMs. A common practice is viewing the language model as a text encoder for extracting semantics. In the beginning, CLIP~\cite{clip} first demonstrates the text-image alignment, and is therefore very popular in image-aligned semantic modeling among various text-to-image generation models~\cite{dalle, rombach2022high, sdxl, tao2023galip}.
With the advent of the \tfive series which are pretrained on text-only corpora, Imagen~\cite{saharia2022photorealistic} observes that \tfive is effective at encoding text for image synthesis. Many works~\cite{pixart_alpha, chen2024pixart, betker2023improving, esser2024scaling, chen2024pixart2} adopt the \tfive series as the text encoding model.
Recently, considering the superior text comprehension capabilities of decoder-only LLMs~\cite{touvron2023llama, llama2, yang2023baichuan, bai2023qwen, young2024yi, team2023internlm, team2023gemini}, some works~\cite{gao2024lumina, zhao2024bridging, hu2024ella, wu2023paragraph, ma2024exploring} try to introduce LLMs into the designed framework. On one hand, LLM2Vec~\cite{behnamghader2024llm2vec} discovers the potential for decoder-only methods to outperform encoder-only methods in both word-level and sequence-level tasks in an unsupervised manner. One the other hand, LiDiT~\cite{ma2024exploring} and SANA~\cite{xie2024sana} provide LLM with complex instructions to encode the prompt for the semantic embedding and train a DiT from scratch for image generation. 
Although ParaDiffusion~\cite{wu2023paragraph} and LaVi-Bridge~\cite{zhao2024bridging} introduce an adapter to bridge \llama and a pretrained generative vision models (\ie, PixArt~\cite{pixart_alpha}), 
we found that the simple adapter does not perform well in video generation due to the complexity of temporal modeling.
Therefore, to the best of our knowledge, our \method is the first work to integrate \llama into the video diffusion framework.
The core of \method is the proposed token fuser which stabilizes fluctuating text features and achieves non-destructive integration of heterogeneous (i.e., encoder and decoder-only) LLMs. Such a design allows the T2V model to fully leverage learned video priors while capitalizing on the text-related capability of LLMs.

\section{Conclusion}

In this paper, we propose a text-to-video diffusion model, \method, which leverages large language model embeddings within the video diffusion transformer to achieve precise text understanding for video spatiotemporal semantics. The core innovation of our approach lies in the token fuser, which fuse semantic features from encoder and decoder-only language models with different distributions. Ablation studies and visualizations validate the effectiveness of \method. Extensive quantitative and qualitative comparisons, along with a detailed user study, demonstrate the superior performance of our method. Please refer to \textit{Supplementary Materials} to view the limitations, ethical considerations and other details.

\renewcommand{\thefootnote}{\arabic{footnote}}
\nopagebreak
{
    \small
    \bibliographystyle{ieeenat_fullname}
    \bibliography{main}

\begin{thebibliography}{60}
\providecommand{\natexlab}[1]{#1}
\providecommand{\url}[1]{\texttt{#1}}
\expandafter\ifx\csname urlstyle\endcsname\relax
  \providecommand{\doi}[1]{doi: #1}\else
  \providecommand{\doi}{doi: \begingroup \urlstyle{rm}\Url}\fi

\bibitem[Abdin et~al.(2024)Abdin, Jacobs, Awan, Aneja, Awadallah, Awadalla, Bach, Bahree, Bakhtiari, Behl, et~al.]{phi3}
Marah Abdin, Sam~Ade Jacobs, Ammar~Ahmad Awan, Jyoti Aneja, Ahmed Awadallah, Hany Awadalla, Nguyen Bach, Amit Bahree, Arash Bakhtiari, Harkirat Behl, et~al.
\newblock Phi-3 technical report: A highly capable language model locally on your phone.
\newblock \emph{arXiv preprint arXiv:2404.14219}, 2024.

\bibitem[Bai et~al.(2023)Bai, Bai, Chu, Cui, Dang, Deng, Fan, Ge, Han, Huang, et~al.]{bai2023qwen}
Jinze Bai, Shuai Bai, Yunfei Chu, Zeyu Cui, Kai Dang, Xiaodong Deng, Yang Fan, Wenbin Ge, Yu Han, Fei Huang, et~al.
\newblock Qwen technical report.
\newblock \emph{arXiv preprint arXiv:2309.16609}, 2023.

\bibitem[Balaji et~al.(2022)Balaji, Nah, Huang, Vahdat, Song, Zhang, Kreis, Aittala, Aila, Laine, et~al.]{balaji2022ediff}
Yogesh Balaji, Seungjun Nah, Xun Huang, Arash Vahdat, Jiaming Song, Qinsheng Zhang, Karsten Kreis, Miika Aittala, Timo Aila, Samuli Laine, et~al.
\newblock ediff-i: Text-to-image diffusion models with an ensemble of expert denoisers.
\newblock \emph{arXiv preprint arXiv:2211.01324}, 2022.

\bibitem[BehnamGhader et~al.(2024)BehnamGhader, Adlakha, Mosbach, Bahdanau, Chapados, and Reddy]{behnamghader2024llm2vec}
Parishad BehnamGhader, Vaibhav Adlakha, Marius Mosbach, Dzmitry Bahdanau, Nicolas Chapados, and Siva Reddy.
\newblock Llm2vec: Large language models are secretly powerful text encoders.
\newblock \emph{arXiv preprint arXiv:2404.05961}, 2024.

\bibitem[Betker et~al.(2023)Betker, Goh, Jing, Brooks, Wang, Li, Ouyang, Zhuang, Lee, Guo, et~al.]{betker2023improving}
James Betker, Gabriel Goh, Li Jing, Tim Brooks, Jianfeng Wang, Linjie Li, Long Ouyang, Juntang Zhuang, Joyce Lee, Yufei Guo, et~al.
\newblock Improving image generation with better captions.
\newblock \emph{Computer Science. https://cdn. openai. com/papers/dall-e-3. pdf}, 2\penalty0 (3):\penalty0 8, 2023.

\bibitem[Blattmann et~al.(2023)Blattmann, Dockhorn, Kulal, Mendelevitch, Kilian, Lorenz, Levi, English, Voleti, Letts, et~al.]{blattmann2023stable}
Andreas Blattmann, Tim Dockhorn, Sumith Kulal, Daniel Mendelevitch, Maciej Kilian, Dominik Lorenz, Yam Levi, Zion English, Vikram Voleti, Adam Letts, et~al.
\newblock Stable video diffusion: Scaling latent video diffusion models to large datasets.
\newblock \emph{arXiv preprint arXiv:2311.15127}, 2023.

\bibitem[Chen et~al.(2023)Chen, Yu, Ge, Yao, Xie, Wu, Wang, Kwok, Luo, Lu, and Li]{pixart_alpha}
Junsong Chen, Jincheng Yu, Chongjian Ge, Lewei Yao, Enze Xie, Yue Wu, Zhongdao Wang, James~T. Kwok, Ping Luo, Huchuan Lu, and Zhenguo Li.
\newblock Pixart-{\(\alpha\)}: Fast training of diffusion transformer for photorealistic text-to-image synthesis.
\newblock \emph{ArXiv}, abs/2310.00426, 2023.

\bibitem[Chen et~al.(2024{\natexlab{a}})Chen, Ge, Xie, Wu, Yao, Ren, Wang, Luo, Lu, and Li]{chen2024pixart}
Junsong Chen, Chongjian Ge, Enze Xie, Yue Wu, Lewei Yao, Xiaozhe Ren, Zhongdao Wang, Ping Luo, Huchuan Lu, and Zhenguo Li.
\newblock Pixart-$\backslash$sigma: Weak-to-strong training of diffusion transformer for 4k text-to-image generation.
\newblock \emph{arXiv preprint arXiv:2403.04692}, 2024{\natexlab{a}}.

\bibitem[Chen et~al.(2024{\natexlab{b}})Chen, Wu, Luo, Xie, Paul, Luo, Zhao, and Li]{chen2024pixart2}
Junsong Chen, Yue Wu, Simian Luo, Enze Xie, Sayak Paul, Ping Luo, Hang Zhao, and Zhenguo Li.
\newblock Pixart-$\{$$\backslash$delta$\}$: Fast and controllable image generation with latent consistency models.
\newblock \emph{arXiv preprint arXiv:2401.05252}, 2024{\natexlab{b}}.

\bibitem[Chen et~al.(2024{\natexlab{c}})Chen, Li, Lin, Zhu, Wang, Yuan, Zhou, Cheng, and Yuan]{chen2024od}
Liuhan Chen, Zongjian Li, Bin Lin, Bin Zhu, Qian Wang, Shenghai Yuan, Xing Zhou, Xinghua Cheng, and Li Yuan.
\newblock Od-vae: An omni-dimensional video compressor for improving latent video diffusion model.
\newblock \emph{arXiv preprint arXiv:2409.01199}, 2024{\natexlab{c}}.

\bibitem[Esser et~al.(2021)Esser, Rombach, and Ommer]{esser2021taming}
Patrick Esser, Robin Rombach, and Bjorn Ommer.
\newblock Taming transformers for high-resolution image synthesis.
\newblock In \emph{Proceedings of the IEEE/CVF conference on computer vision and pattern recognition}, pages 12873--12883, 2021.

\bibitem[Esser et~al.(2024)Esser, Kulal, Blattmann, Entezari, M{\"u}ller, Saini, Levi, Lorenz, Sauer, Boesel, et~al.]{esser2024scaling}
Patrick Esser, Sumith Kulal, Andreas Blattmann, Rahim Entezari, Jonas M{\"u}ller, Harry Saini, Yam Levi, Dominik Lorenz, Axel Sauer, Frederic Boesel, et~al.
\newblock Scaling rectified flow transformers for high-resolution image synthesis.
\newblock In \emph{Forty-first International Conference on Machine Learning}, 2024.

\bibitem[Fang et~al.(2020)Fang, Zhu, Zeng, Ma, and Wang]{fang2020perceptual}
Yuming Fang, Hanwei Zhu, Yan Zeng, Kede Ma, and Zhou Wang.
\newblock Perceptual quality assessment of smartphone photography.
\newblock In \emph{Proceedings of the IEEE/CVF conference on computer vision and pattern recognition}, pages 3677--3686, 2020.

\bibitem[Feng et~al.(2024)Feng, Gong, Chen, Shen, Liu, and Zhou]{feng2024ranni}
Yutong Feng, Biao Gong, Di Chen, Yujun Shen, Yu Liu, and Jingren Zhou.
\newblock Ranni: Taming text-to-image diffusion for accurate instruction following.
\newblock In \emph{Proceedings of the IEEE/CVF Conference on Computer Vision and Pattern Recognition}, pages 4744--4753, 2024.

\bibitem[Gao et~al.(2024)Gao, Zhuo, Lin, Liu, Chen, Du, Xie, Luo, Qiu, Zhang, et~al.]{gao2024lumina}
Peng Gao, Le Zhuo, Ziyi Lin, Chris Liu, Junsong Chen, Ruoyi Du, Enze Xie, Xu Luo, Longtian Qiu, Yuhang Zhang, et~al.
\newblock Lumina-t2x: Transforming text into any modality, resolution, and duration via flow-based large diffusion transformers.
\newblock \emph{arXiv preprint arXiv:2405.05945}, 2024.

\bibitem[Ge et~al.(2023)Ge, Nah, Liu, Poon, Tao, Catanzaro, Jacobs, Huang, Liu, and Balaji]{ge2023preserve}
Songwei Ge, Seungjun Nah, Guilin Liu, Tyler Poon, Andrew Tao, Bryan Catanzaro, David Jacobs, Jia-Bin Huang, Ming-Yu Liu, and Yogesh Balaji.
\newblock Preserve your own correlation: A noise prior for video diffusion models.
\newblock In \emph{Proceedings of the IEEE/CVF International Conference on Computer Vision}, pages 22930--22941, 2023.

\bibitem[Gong et~al.(2024)Gong, Huang, Feng, Zhang, Li, and Liu]{gong2024check}
Biao Gong, Siteng Huang, Yutong Feng, Shiwei Zhang, Yuyuan Li, and Yu Liu.
\newblock Check locate rectify: A training-free layout calibration system for text-to-image generation.
\newblock In \emph{Proceedings of the IEEE/CVF Conference on Computer Vision and Pattern Recognition}, pages 6624--6634, 2024.

\bibitem[Guo et~al.(2023)Guo, Yang, Rao, Liang, Wang, Qiao, Agrawala, Lin, and Dai]{guo2023animatediff}
Yuwei Guo, Ceyuan Yang, Anyi Rao, Zhengyang Liang, Yaohui Wang, Yu Qiao, Maneesh Agrawala, Dahua Lin, and Bo Dai.
\newblock Animatediff: Animate your personalized text-to-image diffusion models without specific tuning.
\newblock \emph{arXiv preprint arXiv:2307.04725}, 2023.

\bibitem[Ho et~al.(2020)Ho, Jain, and Abbeel]{ho2020denoising}
Jonathan Ho, Ajay Jain, and Pieter Abbeel.
\newblock Denoising diffusion probabilistic models.
\newblock \emph{Advances in neural information processing systems}, 33:\penalty0 6840--6851, 2020.

\bibitem[Ho et~al.(2022)Ho, Salimans, Gritsenko, Chan, Norouzi, and Fleet]{ho2022video}
Jonathan Ho, Tim Salimans, Alexey Gritsenko, William Chan, Mohammad Norouzi, and David~J Fleet.
\newblock Video diffusion models.
\newblock \emph{Advances in Neural Information Processing Systems}, 35:\penalty0 8633--8646, 2022.

\bibitem[Hong et~al.(2022)Hong, Ding, Zheng, Liu, and Tang]{cogvideo}
Wenyi Hong, Ming Ding, Wendi Zheng, Xinghan Liu, and Jie Tang.
\newblock Cogvideo: Large-scale pretraining for text-to-video generation via transformers.
\newblock \emph{arXiv preprint arXiv:2205.15868}, 2022.

\bibitem[Hong et~al.(2024)Hong, Wang, Ding, Yu, Lv, Wang, Cheng, Huang, Ji, Xue, et~al.]{hong2024cogvlm2}
Wenyi Hong, Weihan Wang, Ming Ding, Wenmeng Yu, Qingsong Lv, Yan Wang, Yean Cheng, Shiyu Huang, Junhui Ji, Zhao Xue, et~al.
\newblock Cogvlm2: Visual language models for image and video understanding.
\newblock \emph{arXiv preprint arXiv:2408.16500}, 2024.

\bibitem[Hu et~al.(2024)Hu, Wang, Fang, Fu, Cheng, and Yu]{hu2024ella}
Xiwei Hu, Rui Wang, Yixiao Fang, Bin Fu, Pei Cheng, and Gang Yu.
\newblock Ella: Equip diffusion models with llm for enhanced semantic alignment.
\newblock \emph{arXiv preprint arXiv:2403.05135}, 2024.

\bibitem[Huang et~al.(2023)Huang, Sun, Xie, Li, and Liu]{huang2023t2i}
Kaiyi Huang, Kaiyue Sun, Enze Xie, Zhenguo Li, and Xihui Liu.
\newblock T2i-compbench: A comprehensive benchmark for open-world compositional text-to-image generation.
\newblock \emph{Advances in Neural Information Processing Systems}, 36:\penalty0 78723--78747, 2023.

\bibitem[Huang et~al.(2024)Huang, He, Yu, Zhang, Si, Jiang, Zhang, Wu, Jin, Chanpaisit, Wang, Chen, Wang, Lin, Qiao, and Liu]{vbench}
Ziqi Huang, Yinan He, Jiashuo Yu, Fan Zhang, Chenyang Si, Yuming Jiang, Yuanhan Zhang, Tianxing Wu, Qingyang Jin, Nattapol Chanpaisit, Yaohui Wang, Xinyuan Chen, Limin Wang, Dahua Lin, Yu Qiao, and Ziwei Liu.
\newblock {VBench}: Comprehensive benchmark suite for video generative models.
\newblock In \emph{Proceedings of the IEEE/CVF Conference on Computer Vision and Pattern Recognition}, 2024.

\bibitem[Ke et~al.(2021)Ke, Wang, Wang, Milanfar, and Yang]{ke2021musiq}
Junjie Ke, Qifei Wang, Yilin Wang, Peyman Milanfar, and Feng Yang.
\newblock Musiq: Multi-scale image quality transformer.
\newblock In \emph{Proceedings of the IEEE/CVF international conference on computer vision}, pages 5148--5157, 2021.

\bibitem[Kondratyuk et~al.(2023)Kondratyuk, Yu, Gu, Lezama, Huang, Hornung, Adam, Akbari, Alon, Birodkar, et~al.]{kondratyuk2023videopoet}
Dan Kondratyuk, Lijun Yu, Xiuye Gu, Jos{\'e} Lezama, Jonathan Huang, Rachel Hornung, Hartwig Adam, Hassan Akbari, Yair Alon, Vighnesh Birodkar, et~al.
\newblock Videopoet: A large language model for zero-shot video generation.
\newblock \emph{arXiv preprint arXiv:2312.14125}, 2023.

\bibitem[Lab and etc.(2024)]{opensoraplan}
PKU-Yuan Lab and Tuzhan~AI etc.
\newblock Open-sora-plan, 2024.

\bibitem[Lin et~al.(2024)Lin, Liu, Li, and Yang]{lin2024common}
Shanchuan Lin, Bingchen Liu, Jiashi Li, and Xiao Yang.
\newblock Common diffusion noise schedules and sample steps are flawed.
\newblock In \emph{Proceedings of the IEEE/CVF winter conference on applications of computer vision}, pages 5404--5411, 2024.

\bibitem[Luo et~al.(2023)Luo, Chen, Zhang, Huang, Wang, Shen, Zhao, Zhou, and Tan]{VideoFusion}
Zhengxiong Luo, Dayou Chen, Yingya Zhang, Yan Huang, Liang Wang, Yujun Shen, Deli Zhao, Jingren Zhou, and Tieniu Tan.
\newblock Videofusion: Decomposed diffusion models for high-quality video generation.
\newblock In \emph{Proceedings of the IEEE/CVF Conference on Computer Vision and Pattern Recognition (CVPR)}, 2023.

\bibitem[Ma et~al.(2024)Ma, Zong, Song, Li, and Liu]{ma2024exploring}
Bingqi Ma, Zhuofan Zong, Guanglu Song, Hongsheng Li, and Yu Liu.
\newblock Exploring the role of large language models in prompt encoding for diffusion models.
\newblock \emph{arXiv preprint arXiv:2406.11831}, 2024.

\bibitem[Podell et~al.(2023)Podell, English, Lacey, Blattmann, Dockhorn, M{\"u}ller, Penna, and Rombach]{sdxl}
Dustin Podell, Zion English, Kyle Lacey, Andreas Blattmann, Tim Dockhorn, Jonas M{\"u}ller, Joe Penna, and Robin Rombach.
\newblock {SDXL: Improving latent diffusion models for high-resolution image synthesis}.
\newblock \emph{arXiv preprint arXiv:2307.01952}, 2023.

\bibitem[Radford et~al.(2021)Radford, Kim, Hallacy, Ramesh, Goh, Agarwal, Sastry, Askell, Mishkin, Clark, et~al.]{clip}
Alec Radford, Jong~Wook Kim, Chris Hallacy, Aditya Ramesh, Gabriel Goh, Sandhini Agarwal, Girish Sastry, Amanda Askell, Pamela Mishkin, Jack Clark, et~al.
\newblock Learning transferable visual models from natural language supervision.
\newblock In \emph{ICML}, pages 8748--8763, 2021.

\bibitem[Raffel et~al.(2020)Raffel, Shazeer, Roberts, Lee, Narang, Matena, Zhou, Li, and Liu]{t5}
Colin Raffel, Noam Shazeer, Adam Roberts, Katherine Lee, Sharan Narang, Michael Matena, Yanqi Zhou, Wei Li, and Peter~J Liu.
\newblock Exploring the limits of transfer learning with a unified text-to-text transformer.
\newblock \emph{Journal of machine learning research}, 21\penalty0 (140):\penalty0 1--67, 2020.

\bibitem[Ramesh et~al.(2022)Ramesh, Dhariwal, Nichol, Chu, and Chen]{dalle}
Aditya Ramesh, Prafulla Dhariwal, Alex Nichol, Casey Chu, and Mark Chen.
\newblock Hierarchical text-conditional image generation with clip latents.
\newblock \emph{arXiv preprint arXiv:2204.06125}, 1\penalty0 (2):\penalty0 3, 2022.

\bibitem[Rombach et~al.(2022)Rombach, Blattmann, Lorenz, Esser, and Ommer]{rombach2022high}
Robin Rombach, Andreas Blattmann, Dominik Lorenz, Patrick Esser, and Bj{\"o}rn Ommer.
\newblock High-resolution image synthesis with latent diffusion models.
\newblock In \emph{CVPR}, pages 10684--10695, 2022.

\bibitem[Saharia et~al.(2022)Saharia, Chan, Saxena, Li, Whang, Denton, Ghasemipour, Gontijo~Lopes, Karagol~Ayan, Salimans, et~al.]{saharia2022photorealistic}
Chitwan Saharia, William Chan, Saurabh Saxena, Lala Li, Jay Whang, Emily~L Denton, Kamyar Ghasemipour, Raphael Gontijo~Lopes, Burcu Karagol~Ayan, Tim Salimans, et~al.
\newblock Photorealistic text-to-image diffusion models with deep language understanding.
\newblock \emph{NeurIPS}, 35:\penalty0 36479--36494, 2022.

\bibitem[Salimans and Ho(2022)]{salimans2022progressive}
Tim Salimans and Jonathan Ho.
\newblock Progressive distillation for fast sampling of diffusion models.
\newblock \emph{arXiv preprint arXiv:2202.00512}, 2022.

\bibitem[Sohl-Dickstein et~al.(2015)Sohl-Dickstein, Weiss, Maheswaranathan, and Ganguli]{sohl2015deep}
Jascha Sohl-Dickstein, Eric Weiss, Niru Maheswaranathan, and Surya Ganguli.
\newblock Deep unsupervised learning using nonequilibrium thermodynamics.
\newblock In \emph{International conference on machine learning}, pages 2256--2265. PMLR, 2015.

\bibitem[stability.ai(2022)]{stablediffusion2}
stability.ai.
\newblock {Stable Diffusion 2.0 Release}, 2022.

\bibitem[Tao et~al.(2023)Tao, Bao, Tang, and Xu]{tao2023galip}
Ming Tao, Bing-Kun Bao, Hao Tang, and Changsheng Xu.
\newblock Galip: Generative adversarial clips for text-to-image synthesis.
\newblock In \emph{Proceedings of the IEEE/CVF Conference on Computer Vision and Pattern Recognition}, pages 14214--14223, 2023.

\bibitem[Team et~al.(2023)Team, Anil, Borgeaud, Wu, Alayrac, Yu, Soricut, Schalkwyk, Dai, Hauth, et~al.]{team2023gemini}
Gemini Team, Rohan Anil, Sebastian Borgeaud, Yonghui Wu, Jean-Baptiste Alayrac, Jiahui Yu, Radu Soricut, Johan Schalkwyk, Andrew~M Dai, Anja Hauth, et~al.
\newblock Gemini: a family of highly capable multimodal models.
\newblock \emph{arXiv preprint arXiv:2312.11805}, 2023.

\bibitem[Team(2023)]{team2023internlm}
InternLM Team.
\newblock Internlm: A multilingual language model with progressively enhanced capabilities, 2023.

\bibitem[Touvron et~al.(2023{\natexlab{a}})Touvron, Lavril, Izacard, Martinet, Lachaux, Lacroix, Rozi{\`e}re, Goyal, Hambro, Azhar, Rodriguez, Joulin, Grave, and Lample]{touvron2023llama}
Hugo Touvron, Thibaut Lavril, Gautier Izacard, Xavier Martinet, Marie-Anne Lachaux, Timoth{\'e}e Lacroix, Baptiste Rozi{\`e}re, Naman Goyal, Eric Hambro, Faisal Azhar, Aurelien Rodriguez, Armand Joulin, Edouard Grave, and Guillaume Lample.
\newblock Llama: Open and efficient foundation language models.
\newblock \emph{arXiv preprint arXiv:2302.13971}, 2023{\natexlab{a}}.

\bibitem[Touvron et~al.(2023{\natexlab{b}})Touvron, Martin, Stone, Albert, Almahairi, Babaei, Bashlykov, Batra, Bhargava, Bhosale, Bikel, Blecher, Canton{-}Ferrer, Chen, Cucurull, Esiobu, Fernandes, Fu, Fu, Fuller, Gao, Goswami, Goyal, Hartshorn, Hosseini, Hou, Inan, Kardas, Kerkez, Khabsa, Kloumann, Korenev, Koura, Lachaux, Lavril, Lee, Liskovich, Lu, Mao, Martinet, Mihaylov, Mishra, Molybog, Nie, Poulton, Reizenstein, Rungta, Saladi, Schelten, Silva, Smith, Subramanian, Tan, Tang, Taylor, Williams, Kuan, Xu, Yan, Zarov, Zhang, Fan, Kambadur, Narang, Rodriguez, Stojnic, Edunov, and Scialom]{llama2}
Hugo Touvron, Louis Martin, Kevin Stone, Peter Albert, Amjad Almahairi, Yasmine Babaei, Nikolay Bashlykov, Soumya Batra, Prajjwal Bhargava, Shruti Bhosale, Dan Bikel, Lukas Blecher, Cristian Canton{-}Ferrer, Moya Chen, Guillem Cucurull, David Esiobu, Jude Fernandes, Jeremy Fu, Wenyin Fu, Brian Fuller, Cynthia Gao, Vedanuj Goswami, Naman Goyal, Anthony Hartshorn, Saghar Hosseini, Rui Hou, Hakan Inan, Marcin Kardas, Viktor Kerkez, Madian Khabsa, Isabel Kloumann, Artem Korenev, Punit~Singh Koura, Marie{-}Anne Lachaux, Thibaut Lavril, Jenya Lee, Diana Liskovich, Yinghai Lu, Yuning Mao, Xavier Martinet, Todor Mihaylov, Pushkar Mishra, Igor Molybog, Yixin Nie, Andrew Poulton, Jeremy Reizenstein, Rashi Rungta, Kalyan Saladi, Alan Schelten, Ruan Silva, Eric~Michael Smith, Ranjan Subramanian, Xiaoqing~Ellen Tan, Binh Tang, Ross Taylor, Adina Williams, Jian~Xiang Kuan, Puxin Xu, Zheng Yan, Iliyan Zarov, Yuchen Zhang, Angela Fan, Melanie Kambadur, Sharan Narang, Aur{\'{e}}lien Rodriguez, Robert Stojnic, Sergey Edunov,
  and Thomas Scialom.
\newblock Llama 2: Open foundation and fine-tuned chat models.
\newblock \emph{ArXiv}, abs/2307.09288, 2023{\natexlab{b}}.

\bibitem[Van~der Maaten and Hinton(2008)]{van2008visualizing}
Laurens Van~der Maaten and Geoffrey Hinton.
\newblock Visualizing data using t-sne.
\newblock \emph{Journal of machine learning research}, 9\penalty0 (11), 2008.

\bibitem[Wang et~al.(2023{\natexlab{a}})Wang, Lv, Yu, Hong, Qi, Wang, Ji, Yang, Zhao, Song, Xu, Xu, Li, Dong, Ding, and Tang]{wang2023cogvlm}
Weihan Wang, Qingsong Lv, Wenmeng Yu, Wenyi Hong, Ji Qi, Yan Wang, Junhui Ji, Zhuoyi Yang, Lei Zhao, Xixuan Song, Jiazheng Xu, Bin Xu, Juanzi Li, Yuxiao Dong, Ming Ding, and Jie Tang.
\newblock Cogvlm: Visual expert for pretrained language models, 2023{\natexlab{a}}.

\bibitem[Wang et~al.(2023{\natexlab{b}})Wang, He, Li, Li, Yu, Ma, Li, Chen, Chen, Wang, et~al.]{wang2023internvid}
Yi Wang, Yinan He, Yizhuo Li, Kunchang Li, Jiashuo Yu, Xin Ma, Xinhao Li, Guo Chen, Xinyuan Chen, Yaohui Wang, et~al.
\newblock Internvid: A large-scale video-text dataset for multimodal understanding and generation.
\newblock \emph{arXiv preprint arXiv:2307.06942}, 2023{\natexlab{b}}.

\bibitem[Wu et~al.(2025)Wu, Wang, Yang, Gan, Liu, Yuan, and Wang]{wu2025grit}
Jialian Wu, Jianfeng Wang, Zhengyuan Yang, Zhe Gan, Zicheng Liu, Junsong Yuan, and Lijuan Wang.
\newblock Grit: A generative region-to-text transformer for object understanding.
\newblock In \emph{European Conference on Computer Vision}, pages 207--224. Springer, 2025.

\bibitem[Wu et~al.(2023)Wu, Li, He, Shou, Shen, Cheng, Li, Gao, Zhang, and Wang]{wu2023paragraph}
Weijia Wu, Zhuang Li, Yefei He, Mike~Zheng Shou, Chunhua Shen, Lele Cheng, Yan Li, Tingting Gao, Di Zhang, and Zhongyuan Wang.
\newblock Paragraph-to-image generation with information-enriched diffusion model.
\newblock \emph{arXiv preprint arXiv:2311.14284}, 2023.

\bibitem[Xie et~al.(2024)Xie, Chen, Chen, Cai, Lin, Zhang, Li, Lu, and Han]{xie2024sana}
Enze Xie, Junsong Chen, Junyu Chen, Han Cai, Yujun Lin, Zhekai Zhang, Muyang Li, Yao Lu, and Song Han.
\newblock Sana: Efficient high-resolution image synthesis with linear diffusion transformers.
\newblock \emph{arXiv preprint arXiv:2410.10629}, 2024.

\bibitem[Yang et~al.(2023)Yang, Xiao, Wang, Zhang, Bian, Yin, Lv, Pan, Wang, Yan, et~al.]{yang2023baichuan}
Aiyuan Yang, Bin Xiao, Bingning Wang, Borong Zhang, Ce Bian, Chao Yin, Chenxu Lv, Da Pan, Dian Wang, Dong Yan, et~al.
\newblock Baichuan 2: Open large-scale language models.
\newblock \emph{arXiv preprint arXiv:2309.10305}, 2023.

\bibitem[Yang et~al.(2024)Yang, Teng, Zheng, Ding, Huang, Xu, Yang, Hong, Zhang, Feng, et~al.]{cogvideox}
Zhuoyi Yang, Jiayan Teng, Wendi Zheng, Ming Ding, Shiyu Huang, Jiazheng Xu, Yuanming Yang, Wenyi Hong, Xiaohan Zhang, Guanyu Feng, et~al.
\newblock Cogvideox: Text-to-video diffusion models with an expert transformer.
\newblock \emph{arXiv preprint arXiv:2408.06072}, 2024.

\bibitem[Young et~al.(2024)Young, Chen, Li, Huang, Zhang, Zhang, Li, Zhu, Chen, Chang, et~al.]{young2024yi}
Alex Young, Bei Chen, Chao Li, Chengen Huang, Ge Zhang, Guanwei Zhang, Heng Li, Jiangcheng Zhu, Jianqun Chen, Jing Chang, et~al.
\newblock Yi: Open foundation models by 01. ai.
\newblock \emph{arXiv preprint arXiv:2403.04652}, 2024.

\bibitem[Yu et~al.(2023{\natexlab{a}})Yu, Cheng, Sohn, Lezama, Zhang, Chang, Hauptmann, Yang, Hao, Essa, et~al.]{yu2023magvit}
Lijun Yu, Yong Cheng, Kihyuk Sohn, Jos{\'e} Lezama, Han Zhang, Huiwen Chang, Alexander~G Hauptmann, Ming-Hsuan Yang, Yuan Hao, Irfan Essa, et~al.
\newblock Magvit: Masked generative video transformer.
\newblock In \emph{Proceedings of the IEEE/CVF Conference on Computer Vision and Pattern Recognition}, pages 10459--10469, 2023{\natexlab{a}}.

\bibitem[Yu et~al.(2023{\natexlab{b}})Yu, Lezama, Gundavarapu, Versari, Sohn, Minnen, Cheng, Gupta, Gu, Hauptmann, et~al.]{yu2023language}
Lijun Yu, Jos{\'e} Lezama, Nitesh~B Gundavarapu, Luca Versari, Kihyuk Sohn, David Minnen, Yong Cheng, Agrim Gupta, Xiuye Gu, Alexander~G Hauptmann, et~al.
\newblock Language model beats diffusion--tokenizer is key to visual generation.
\newblock \emph{arXiv preprint arXiv:2310.05737}, 2023{\natexlab{b}}.

\bibitem[Yu et~al.(2024)Yu, Cheng, Wang, Kumar, Macherey, Huang, Ross, Essa, Bisk, Yang, et~al.]{yu2024spae}
Lijun Yu, Yong Cheng, Zhiruo Wang, Vivek Kumar, Wolfgang Macherey, Yanping Huang, David Ross, Irfan Essa, Yonatan Bisk, Ming-Hsuan Yang, et~al.
\newblock Spae: Semantic pyramid autoencoder for multimodal generation with frozen llms.
\newblock \emph{Advances in Neural Information Processing Systems}, 36, 2024.

\bibitem[Zhao et~al.(2024{\natexlab{a}})Zhao, Hao, Zi, Xu, and Wong]{zhao2024bridging}
Shihao Zhao, Shaozhe Hao, Bojia Zi, Huaizhe Xu, and Kwan-Yee~K Wong.
\newblock Bridging different language models and generative vision models for text-to-image generation.
\newblock \emph{arXiv preprint arXiv:2403.07860}, 2024{\natexlab{a}}.

\bibitem[Zhao et~al.(2024{\natexlab{b}})Zhao, Zhang, Cun, Yang, Niu, Li, Hu, and Shan]{zhao2024cv}
Sijie Zhao, Yong Zhang, Xiaodong Cun, Shaoshu Yang, Muyao Niu, Xiaoyu Li, Wenbo Hu, and Ying Shan.
\newblock Cv-vae: A compatible video vae for latent generative video models.
\newblock \emph{arXiv preprint arXiv:2405.20279}, 2024{\natexlab{b}}.

\bibitem[Zheng et~al.(2024)Zheng, Peng, Yang, Shen, Li, Liu, Zhou, Li, and You]{opensora}
Zangwei Zheng, Xiangyu Peng, Tianji Yang, Chenhui Shen, Shenggui Li, Hongxin Liu, Yukun Zhou, Tianyi Li, and Yang You.
\newblock Open-sora: Democratizing efficient video production for all, march 2024.
\newblock \emph{URL https://github. com/hpcaitech/Open-Sora}, 1\penalty0 (3):\penalty0 4, 2024.

\end{thebibliography}
}

\clearpage
\setcounter{page}{1}
\maketitlesupplementary
\appendix

In the main paper, we provide a method diagram and textual description of \method. Here, we present the detailed pseudocode of the \textbf{Token Fuser} in \method in Algorithm~\ref{alg1} for direct reference. In the following sections, We introduce the data processing in Sec.~\ref{sup:dp}, the evaluation metric in Sec.~\ref{sec:Evaluation_Metric}, the user study in Sec.~\ref{sup:us}, and additional experimental results in Sec.~\ref{sup:addexp}. We also introduce limitations and social impact of our work in Sec.~\ref{sup:lim} and Sec.~\ref{sup:si} respectively.

\section{Data Processing}
\label{sup:dp}

\begin{figure}[t]
  \centering
    \includegraphics[width=1\linewidth]{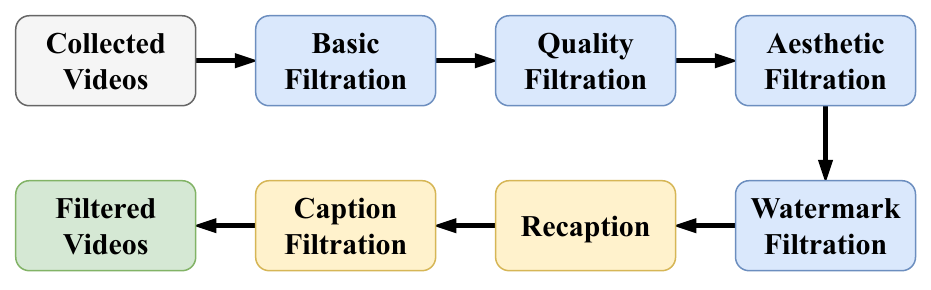}
    \caption{The pipeline for preparing data.}
    \label{fig:data_pipeline}
\end{figure}

We construct a collection of relatively high-quality video clips with text descriptions using a combination of video filtering and recaptioning models. As shown in Fig.~\ref{fig:data_pipeline}, the collected data undergoes multiple filtration steps: Basic Filtration, Quality Filtration, Aesthetic Filtration, Watermark
Filtration, which removes data that does not meet fundamental requirements. After these video-based filtration steps, captions are generated for the videos. The videos and their captions are then evaluated for consistency to ensure the caption accurately describes the video content. Following this process, approximately 500,000 single-shot clips remain, with each clip averaging about 10 seconds. These high-quality video clips are ultimately used for training Mimir. Next, we provide a detailed explanation of each stage of this pipeline.

\begin{figure}[t]

\begin{algorithm}[H]
\caption{\small Token Fuser}
\label{alg:train}
\definecolor{codeblue}{rgb}{0.1,0.6,0.1}
\definecolor{codekw}{rgb}{0.85, 0.18, 0.50}
\lstset{
  backgroundcolor=\color{white},
  basicstyle=\fontsize{7.5pt}{7.5pt}\ttfamily\selectfont,
  columns=fullflexible,
  breaklines=true,
  captionpos=b,
  commentstyle=\fontsize{7.5pt}{7.5pt}\color{codeblue},
  keywordstyle=\fontsize{7.5pt}{7.5pt}\color{codekw},
  escapechar={|}, 
}
\begin{lstlisting}[language=python]
# Inputs

# Text prompt provided by the user
text_prompt = "Input text prompt"          
# Instructional input for fine-tuning
instruction_prompt = "Instruction description"  

# 1. Encoding and Tokenization

# Obtain token embeddings from text encoder
e_theta = TextEncoder(text_prompt)       
# Obtain token embeddings from decoder-only model
e_beta = DecoderModel(text_prompt)              
# Obtain instruction token from decoder-only model
e_i = DecoderModel(instruction_prompt)         

# 2. Token Fusion to Address Feature Distribution Gap
# Normalize e_beta and apply learnable scale
# Apply normalization
e_beta = Normalize(e_beta)    
# Scale normalized features
e_beta = LearnableScale(e_beta) 

# Apply Zero-Conv to e_beta and e_theta to maintain original semantic space
# Maintain semantic space for e_beta
e_beta = ZeroConv(e_beta)          
# Maintain semantic space for e_theta
e_theta = e_theta + ZeroConv(e_theta)               

# Sum modified tokens to form combined tokens
e = e_theta + e_beta  # Shape: [n, 4096]

# 3. Stabilizing Divergent Semantic Features
# Initialize learnable tokens and add to instruction tokens
# Four learnable tokens, shape: [4, 4096]
e_l = InitializeLearnableTokens(count=4, dim=4096)  
# Stabilize instruction features
e_s = e_i + e_l                           

# 4. Final Token Fusion and Video Generation
# Concatenate e_combined and stabilized tokens
e_final = Concatenate(e_combined, e_stabilized)     # Shape: [n+4, 4096]

# Generate videos using the final fused tokens
generated_video = VideoGenerator(e_final)

# Output
return generated_video

\end{lstlisting}
\label{alg1}
\end{algorithm}
\end{figure}

\noindent \textbf{Basic Filtration.} 
At this stage, we focus on computing video metadata and filtering out invalid videos.
\begin{enumerate}
    \item \textit{Metadata Extraction:} Most of important video properties such as length, width, frame rate, frame count, and duration are obtained and saving using FFmpeg.
    \item \textit{Filtering Rules:}
    \begin{itemize}
        \item Videos with fewer than 65 frames, a duration of less than 1s, or an aspect ratio (width / height) outside the range [1, 2] are excluded.
        \item Videos with a motion score of 0, determined using optical flow, are excluded.
    \end{itemize}

\end{enumerate}

\noindent \textbf{Quality Filtration.}
At this stage, we calculate basic quality indicators for the videos and remove those that do not meet the standards.
\begin{enumerate}
    \item \textit{Quality Metrics:} We use OpenCV to calculate the black area percentage, brightness, and black frame rate.
    \item \textit{Filtering Rules:} 
    \begin{itemize}
        \item Black area $>$ 0.8, excluding.
        \item Brightness  $<$ 0.2, excluding.
        \item Black frame rate $>$ 0.4, excluding.
    \end{itemize}
\end{enumerate}

\noindent \textbf{Aesthetic Filtration.} 
At this stage, we filter videos based on aesthetic-related operators.
\begin{enumerate}
    \item \textit{Aesthetic Metrics:} We use the aesthetic predictor~\footnote{https://github.com/christophschuhmann/improved-aesthetic-predictor} to calculate aesthetic score and OCR coverage.
    \item \textit{Filtering Rules: }
    \begin{itemize}
        \item Aesthetic score $<$ 4.0, excluding.
        \item OCR coverage  $>$ 0.1, excluding.
    \end{itemize}

\end{enumerate}

\noindent \textbf{Watermark Filtration.} 
At this stage, videos containing watermarks are excluded. Each video is analyzed using QWen2-VL-7B~\cite{bai2023qwen} to detect the presence of watermarks. Videos flagged as ``\textit{containing watermarks}'' are excluded.

\noindent \textbf{Re-Caption.} 
At this stage, we use CogVim2~\cite{wang2023cogvlm, hong2024cogvlm2} to generate captions, which produces semantic and detailed descriptions of visual contents in videos.

\noindent \textbf{Caption Filtration.} 
Due to hallucinations in large language models, not all output captions are immediately usable. To address this, we employ human designed rule-based methods and text quality metrics to clean the captions.
\begin{enumerate}
    \item \textit{Text Quality Metrics:}
    \begin{itemize}
        \item N-gram~\footnote{https://github.com/EurekaLabsAI/ngram} repetition rates
        \item Semantic alignment between the video and the generated caption using CLIP Score.
    \end{itemize}
    \item \textit{Filtering Rules:}
    \begin{itemize}
        \item 2-gram repetition $>$ 0.056, excluding.
        \item 5-gram repetition $>$ 0.047, excluding.
        \item 10-gram repetition $>$ 0.045, excluding.
        \item Semantic consistency (CLIP score) $<$  0.25, excluding.
    \end{itemize}
\end{enumerate}
This pipeline ensures the collection of high-quality video clips with accurate captions, which are suitable for training.

\section{Evaluation Metric}
\label{sec:Evaluation_Metric}
We employ several evaluation metrics in VBench~\cite{vbench} to quantitatively assess our results, including \textit{Background Consistency}, \textit{Aesthetic Quality}, \textit{Imaging Quality}, \textit{Object Class}, \textit{Multiple Objects}, \textit{Color Consistency}, \textit{Spatial Relationship}, and \textit{Temporal Style}. The detailed metrics are introduced as follows:

\begin{itemize}
    \item \textit{Background Consistency}. This metric evaluates the temporal consistency of background scenes by calculating the similarity of CLIP~\cite{clip} features across frames.
    \item  \textit{Aesthetic Quality}. This assesses the artistic and aesthetic value perceived by humans for each video frame using the LAION aesthetic predictor. It reflects qualities such as layout, color richness and harmony, photo-realism, naturalness, and overall artistic quality across frames.
    \item \textit{Imaging Quality}. This measures distortions (e.g., over-exposure, noise, blur) present in generated frames. It is evaluated using the MUSIQ~\cite{ke2021musiq} image quality predictor trained on the SPAQ~\cite{fang2020perceptual} dataset.
    \item \textit{Object Class}. This metric is computed using GRiT~\cite{wu2025grit} to measure the success rate of generating the specific object classes described in the text prompt.
    \item \textit{Multiple Objects}. This evaluates the success rate of generating all the objects specified in the text prompt within each video frame. Beyond generating a single object, it assesses the model's ability to compose multiple objects from different classes in the same frame, which is an essential aspect of video generation.
    \item \textit{Color Consistency}. This measures whether the synthesized object colors align with the text prompt. It uses GRiT~\cite{wu2025grit} for color captioning and compares the results against the expected color.
    \item \textit{Spatial Relationship}. This metric evaluates whether the spatial relationships in the generated video follow those specified by the text prompt. It focuses on four primary types of spatial relationships and performs rule-based evaluation similar to~\cite{huang2023t2i}.
    \item \textit{Temporal Style}. This assesses the consistency of temporal style by using ViCLIP~\cite{wang2023internvid} to calculate the similarity between video features and temporal features.
\end{itemize}

\section{User Study}
\label{sup:us}
To obtain genuine feedback reflective of practical applications, the 10 participants in our user study experiment come from diverse academic backgrounds. Since many of them do not major in computer vision, we provide detailed explanations for each question to assist their judgments.
\begin{itemize}
\item Instruction Following: Determine which video aligns more closely with the prompt, evaluate whether the main content is adequately presented in the video, and assess the accuracy and completeness of the prompt.
\item Physics Simulation: Determine which video aligns more closely with real-world physical laws, including object motion, transformations, and other dynamics.
\item Visual Quality: Determine which video has a more harmonious overall visual composition and showcases finer details more exquisitely.

\end{itemize}

\section{Additional Experimental Results}
\label{sup:addexp}
\subsection{Short / Long Prompt}
To investigate the performance differences of \method when inputting short and coarse prompts versus long and fine

\begin{figure*}[!t]
  \centering
    \includegraphics[width=0.95\linewidth]{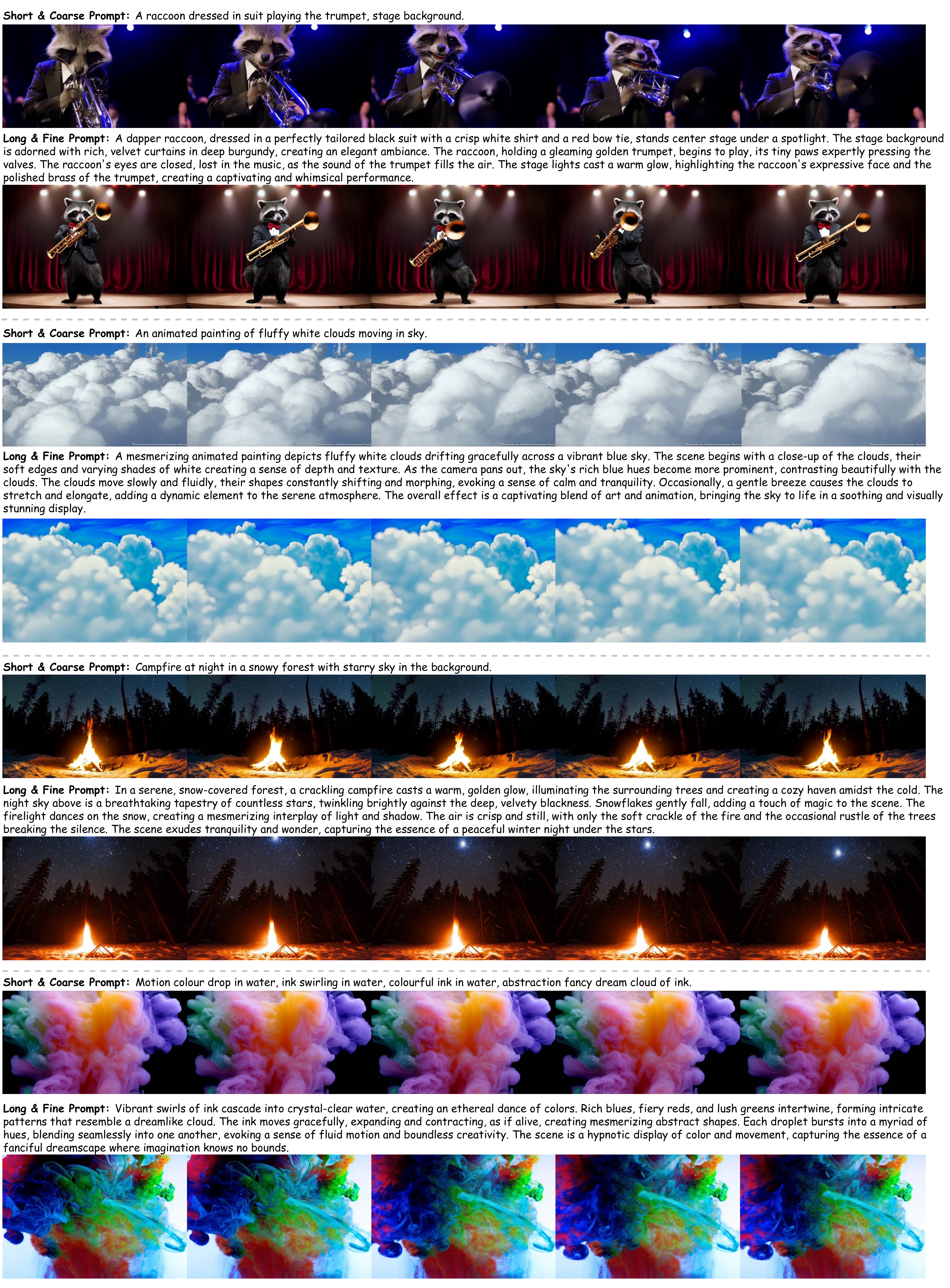}
    \vspace{-4mm}
    \caption{The comparison between results with short \& course prompts and long \& fine prompts.}
    \label{fig:short_long}
    \vspace{-10mm}
\end{figure*}

\begin{figure*}[!htbp]
  \centering
    \vspace{3mm}
    \includegraphics[width=1\linewidth]{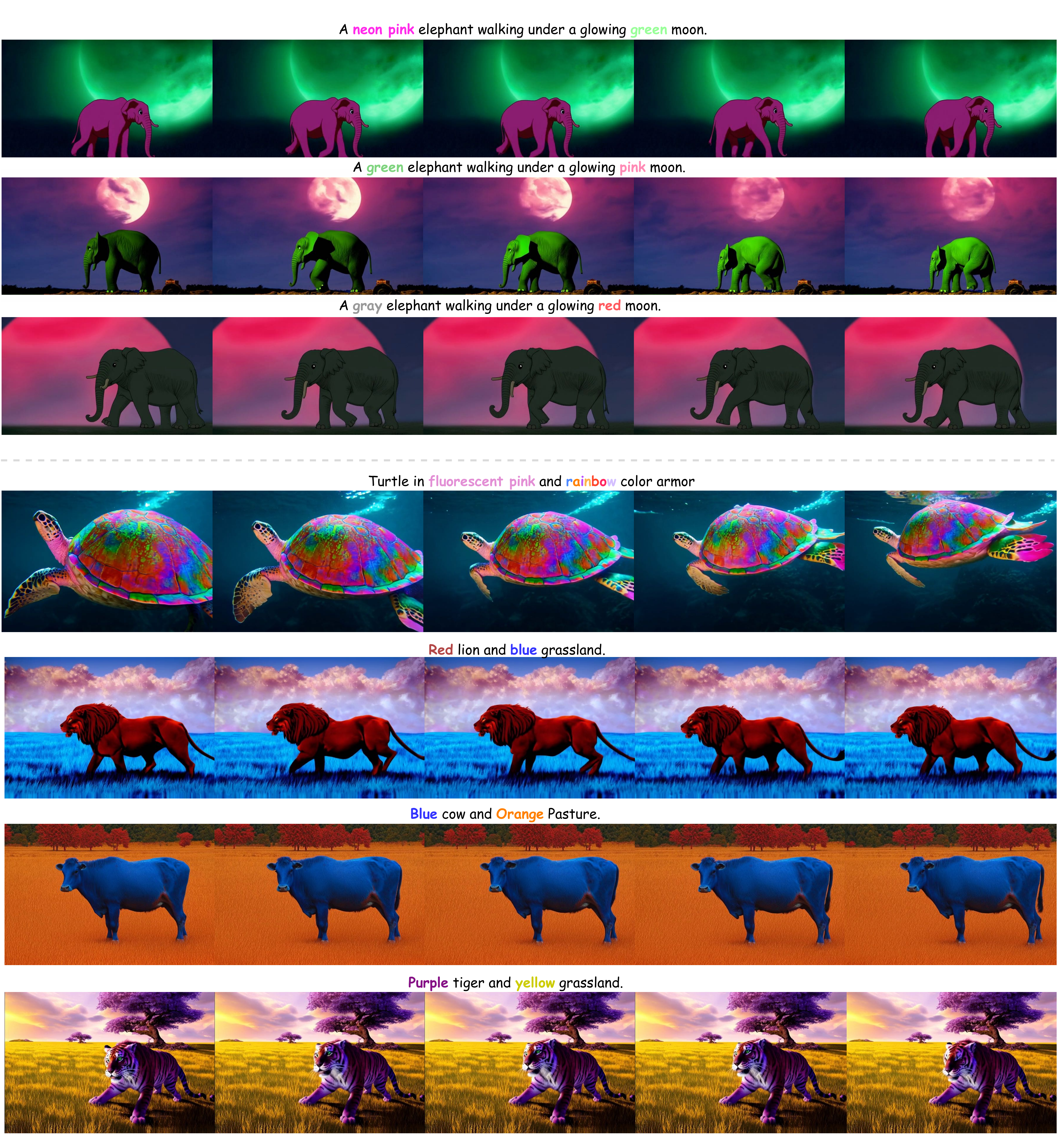}
    \vspace{-8mm}
    \caption{More examples in terms of color rendering.}
    \label{fig:more_color_case}
    \vspace{-3mm}
\end{figure*}

\noindent prompts, we randomly sampled 4 prompts from the VBench dataset. Additionally, VBench provides enhanced versions of these 4 prompts through a large language model. We input both versions into \method and generated corresponding videos. As shown in Fig.~\ref{fig:short_long}, leveraging the reasoning ability of the decoder-only LLM, even with short and coarse prompts, \method can generate results as detailed as those produced with long and fine prompts. This demonstrates that \method's token fuser effectively expands the semantic, leading to precise text understanding capabilities.

\begin{figure*}[!htbp]
  \centering
    \includegraphics[width=1\linewidth]{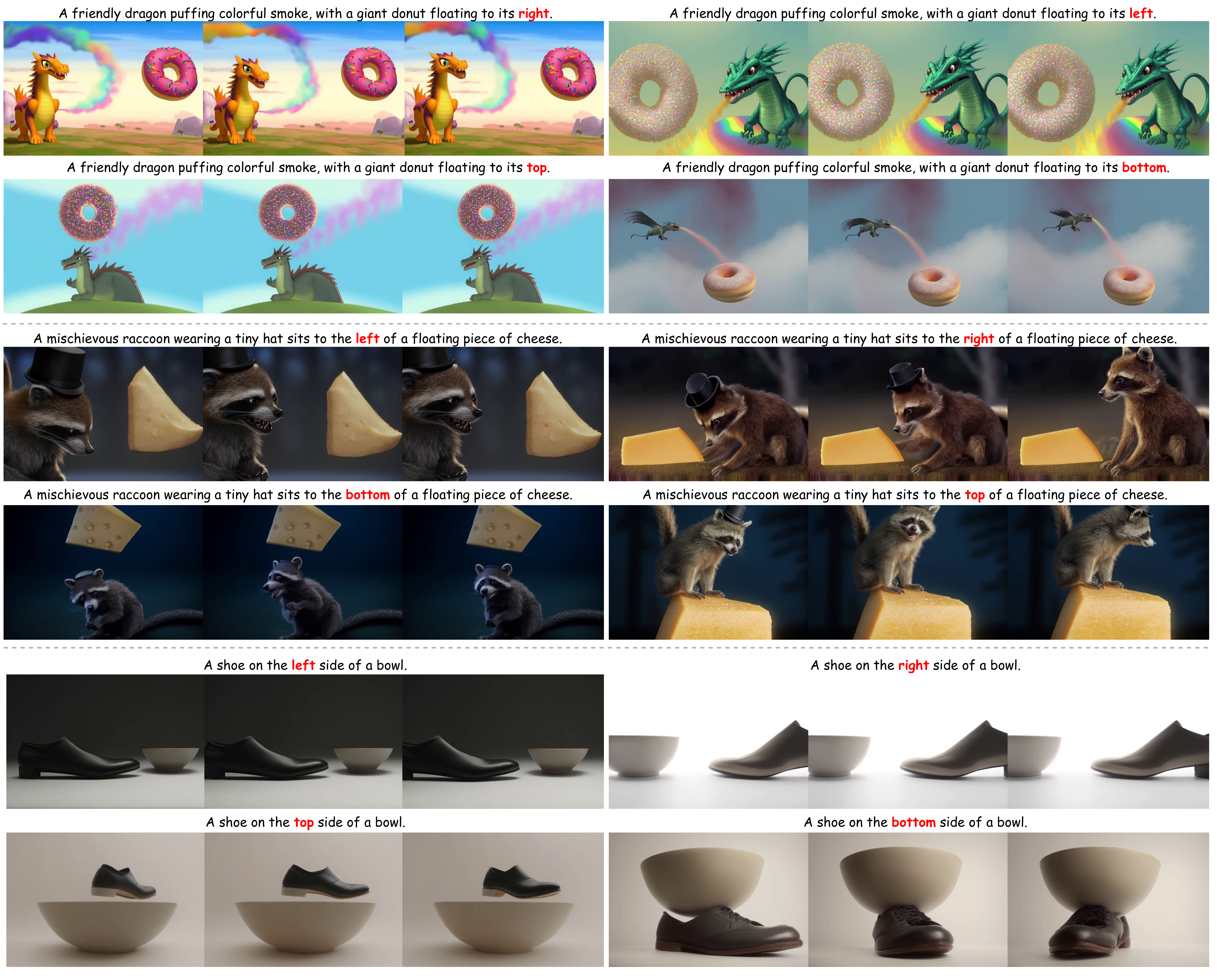}
    \vspace{-8mm}
    \caption{More examples in terms of absolute \& relative position.}
    \label{fig:more_spatial_case}
    \vspace{-5mm}
\end{figure*}

\subsection{More Interesting Prompts}

\subsubsection{Spatial Semantic Understanding}
\noindent \textbf{Color Rendering.} 
As shown in Fig.~\ref{fig:more_color_case}, our method demonstrates the ability to accurately understand the color specifications in the prompt for different objects and generates videos containing objects with the correct colors. It highlights the effectiveness of our token fuser in ensuring semantic alignment between the input prompt and the generated video. By accurately capturing and representing color details, \method  delivers coherent results, even in cases where multiple objects with distinct colors are specified.

\noindent \textbf{Absolute \& Relative Position.} 
As shown in Fig.~\ref{fig:more_spatial_case}, our method effectively understands the spatial relationships (\textit{i.e.}, the absolute \& relative position) specified in the prompt, such as ``\textit{top}'', ``\textit{below}'', ``\textit{left}'', and ``\textit{right}'' and generates videos where objects are positioned correctly according to these relationships. By accurately representing spatial arrangements, \method ensures that the generated videos meet the semantic requirements of complex prompts involving positional relationships between objects.

\begin{figure*}[!htbp]
  \centering
    \includegraphics[width=1\linewidth]{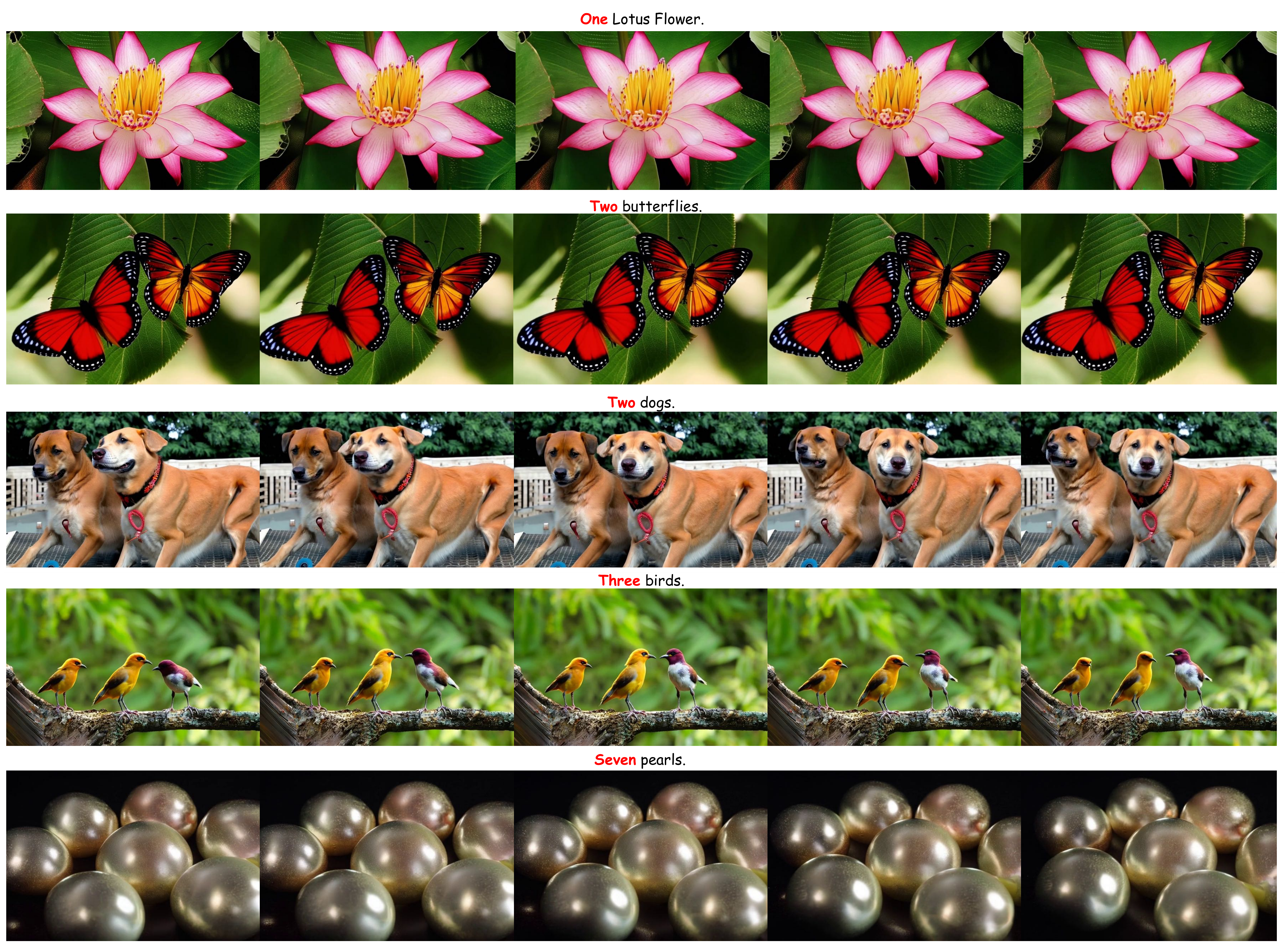}
    \vspace{-8mm}
    \caption{More examples in terms of counting.}
    \label{fig:more_num_case}
\end{figure*}

\noindent \textbf{Counting.} 
As shown in Fig.~\ref{fig:more_num_case}, \method demonstrates a strong ability to understand counting. For example, if the prompt specifies a certain number of objects, \method accurately interprets this information and generates videos containing the correct quantity. By successfully handling quantity-specific prompts, \method proves its reliability in scenarios where precise numeric understanding is critical for video generation tasks.

\begin{figure*}[!htbp]
  \centering
    \includegraphics[width=1\linewidth]{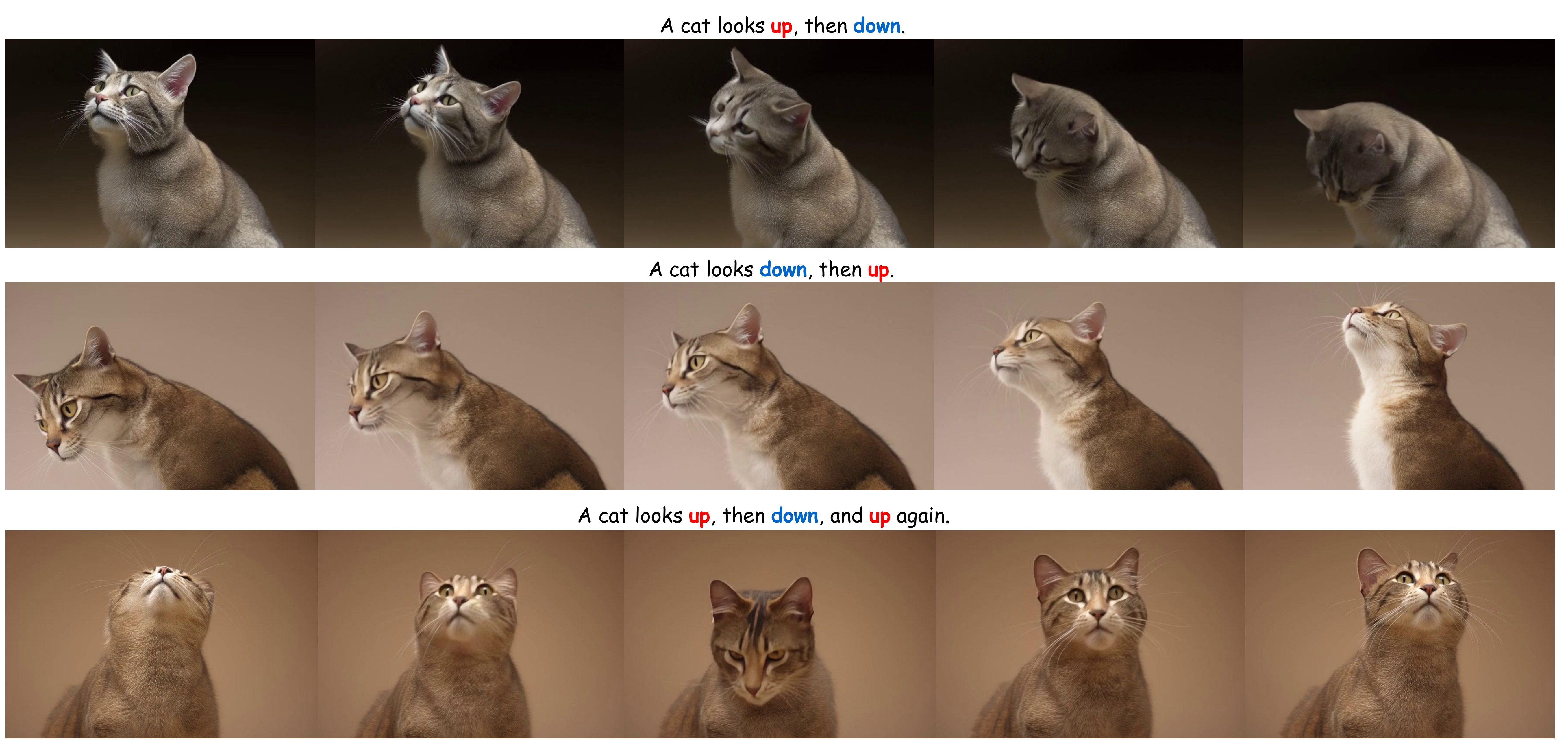}
    \vspace{-8mm}
    \caption{More examples in terms of action sequence over time.}
    \label{fig:more_shunxu_cases}
    \vspace{-8mm}
\end{figure*}

\subsubsection{Temporal Semantic Understanding}
\noindent \textbf{Sequential Actions.}  This involves capturing the sequence of actions performed by an object, such as a cat looking up, then down, or following a more complex pattern like up, down, and up again. It requires precise temporal understanding to maintain the correct order of actions. As shown in Fig.~\ref{fig:more_shunxu_cases}, \method precisely interprets and reproduces these action sequences.

\begin{figure*}[!htbp]
  \centering
    \includegraphics[width=1\linewidth]{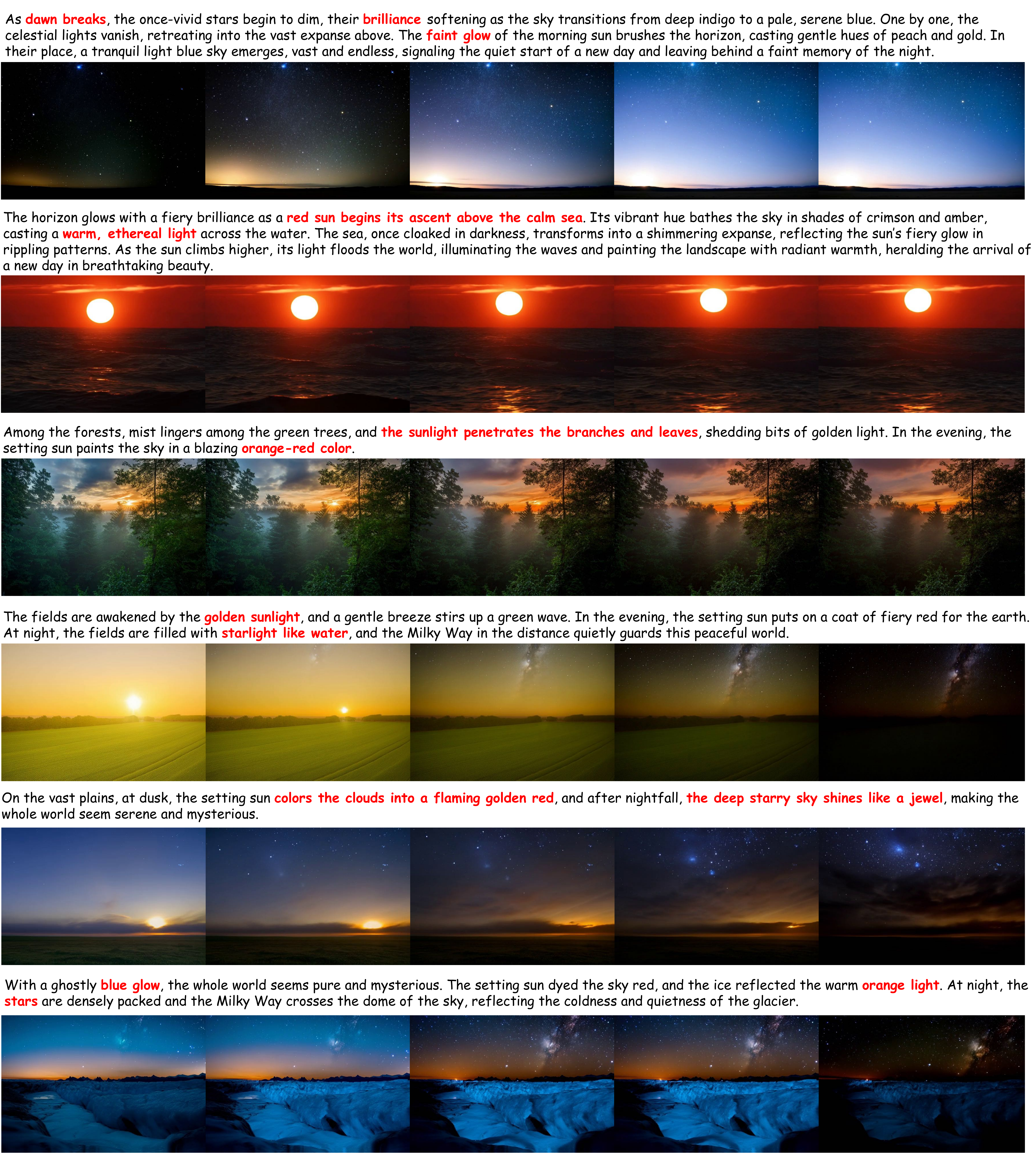}
    \vspace{-8mm}
    \caption{More examples in terms of light changes, showcasing the illumination harmonization over time.}
    \label{fig:more_time_sky_cases}
    \vspace{-3mm}
\end{figure*}

\noindent \textbf{Illumination Harmonization.} It means light changes in the environment, such as dawn transitioning to sunrise and then to sunset. As shown in Fig.~\ref{fig:more_time_sky_cases}, \method precisely generates these gradual scene changes, ensuring the illumination harmonization and the alignment with prompts.

\begin{figure*}[!htbp]
  \centering
    \includegraphics[width=1\linewidth]{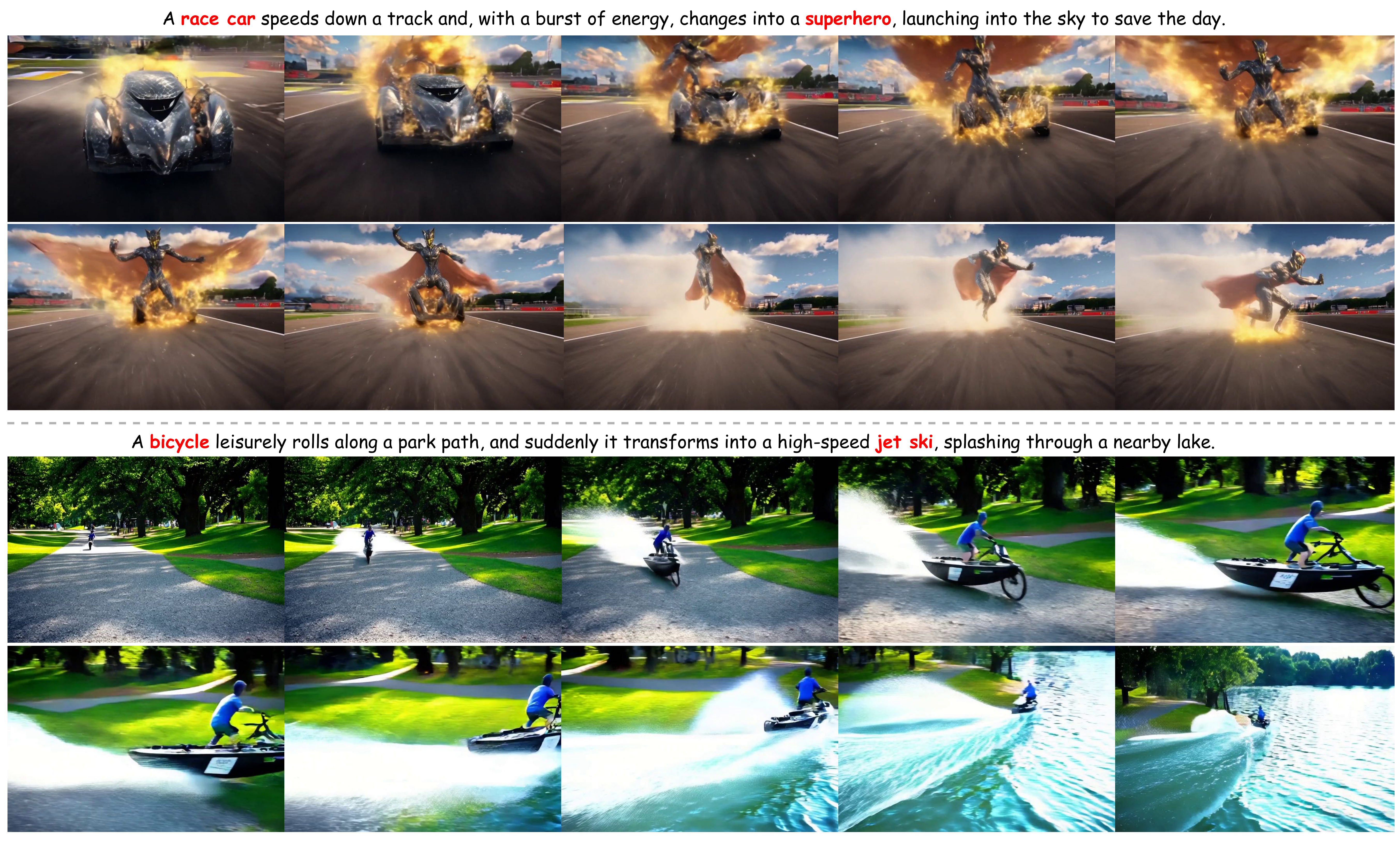}
    \vspace{-8mm}
    \caption{More examples in terms of object transformation over time.}
    \label{fig:more_timechange_cases}
    \vspace{-3mm}
\end{figure*}

\noindent \textbf{Object Transformation.} 
It means transforming an object into another, such as a car transforming into a superhero. This is a highly challenging task due to the complexity of capturing smooth transitions. As shown in Fig.~\ref{fig:more_timechange_cases}, \method precisely understands the prompt and generates well.

\section{Limitations}
\label{sup:lim}
While our current work has made significant strides, it also possesses certain limitations. Firstly, the generated videos are typically limited to short durations (a few seconds to tens of seconds). This is primarily due to the significant computational resources and storage requirements needed for generating longer videos. Additionally, extending the video length may exacerbate temporal inconsistencies, such as discontinuities in actions or backgrounds across frames, which can detract from the overall quality and realism. Secondly, the effectiveness of our T2V model is heavily dependent on the quality and diversity of the training data. In domains where the training dataset lacks coverage—such as specific professional scenarios—the model's performance can be suboptimal. This limitation highlights the importance of expanding and diversifying training datasets to improve the model's generalizability across a broader range of applications.

\section{Social Impact}
\label{sup:si}
Our proposed T2V (Text-to-Video) model demonstrates strong potential for generating high-quality, contextually accurate video content directly from textual descriptions. This technology offers significant benefits across various domains, enabling more accessible, creative, and automated video generation workflows. However, like any generative technology, our T2V model also raises concerns about potential misuse. Malicious actors could exploit it to produce deceptive or harmful video content, such as fake news or misleading advertisements, amplifying the spread of misinformation on social media platforms. This misuse could lead to detrimental societal consequences, including the erosion of trust in digital media. Despite ongoing advancements in generative content detection technologies, challenges remain, especially in scenarios involving complex, high-quality synthetic videos. To address this, we are committed to promoting responsible use of T2V technology and actively contributing to the research community. We aim to share our generated results to support the development of more robust detection algorithms, fostering a safer digital environment capable of mitigating the risks associated with increasingly sophisticated generative models.

\end{document}